\documentclass[a4paper,fleqn, review]{cas-dc}

\usepackage[numbers]{natbib}

\def\tsc#1{\csdef{#1}{\textsc{\lowercase{#1}}\xspace}}
\tsc{WGM}
\tsc{QE}
\tsc{EP}
\tsc{PMS}
\tsc{BEC}
\tsc{DE}

\usepackage{amsmath,amsfonts, nccmath}
\usepackage{graphicx}
\usepackage{textcomp}
\usepackage{xcolor}
\usepackage{listings}
\usepackage{algorithm}% http://ctan.org/pkg/algorithms
\usepackage{algpseudocode}% \usepackage{hyperref}
\usepackage{verbatim}
\usepackage{subcaption}
\usepackage{hyperref}
\usepackage{caption}
\usepackage{amsmath}
\usepackage{balance}
\usepackage{tcolorbox}
\usepackage[dvipsnames]{xcolor}
\begin{document}

\newcommand{\Foutse}[1]{\textcolor{red}{{\it [Foutse: #1]}}}
\newcommand{\giulio}[1]{\textcolor{blue}{{\it [Giulio says: #1]}}}
\newcommand{\Dima}[1]{\textcolor{black}{{ #1  }}}

%\title [mode = title]{Automatic Search-Based  Scenario Generation for Cyber-Physical Systems}  
%\title [mode = title]{Search-Based Virtual Environment Generation for Autonomous Cyber-Physical Systems Testing \Foutse{the title should make it clear that it is a work about testing ...search engines use titles to classify papers! so when picking a title, make sure key concepts are reflected!}}  
\title[mode = title]{A Search-Based Framework for Automatic Generation of Testing Environments for Cyber-Physical Systems}
%\author{Anonymous authors}
\author{Dmytro Humeniuk}\cormark[1]
\author {Foutse Khomh}
\author {Giuliano Antoniol}

\address{Polytechnique Montréal, 2500 Chemin de Polytechnique,  QC H3T 1J4, Montréal, Canada }
\cortext[cor1]{Corresponding author, e-mail: dmytro.humeniuk@polymtl.ca}
%\author{\IEEEauthorblockN{Dmytro Humeniuk}
%\IEEEauthorblockA{\textit{Polytechnique Montréal} \\
%Montreal, Canada \\
%dmytro.humeniuk@polymtl.ca
%}
%\and
%\IEEEauthorblockN{Giuliano Antoniol}
%\IEEEauthorblockA{\textit{Polytechnique Montréal} \\
%Montreal, Canada \\
%giuliano.antoniol@polymtl.ca}
%\and
%\IEEEauthorblockN{Foutse Khomh}
%\IEEEauthorblockA{\textit{Polytechnique Montréal} \\
%Montreal, Canada \\
%foutse.khomh@polymtl.ca}
%}
\begin{abstract}
\textbf{\textbf{Background.}} Many modern cyber-physical systems incorporate computer vision technologies, complex sensors and advanced control software, allowing them to interact with the environment autonomously. Examples include drone swarms, self-driving vehicles, autonomous robots, etc. Testing such systems poses numerous challenges: not only should the system inputs be varied, but also the surrounding environment should be accounted for. A number of tools have been developed to test the system model for the possible inputs falsifying its requirements. However, they are not directly applicable to autonomous cyber-physical systems, as the inputs to their models are generated while operating in a virtual environment.\\
\textbf{Aims.} In this paper, we aim to design a search-based framework, named AmbieGen, for generating diverse fault-revealing test scenarios for autonomous cyber-physical systems. The scenarios represent an environment in which an autonomous agent operates. The framework should be applicable to generating different types of environments.
%The test cases should account for varying environmental conditions and surroundings, and the algorithm should be applicable to different.
\\
\textbf{Method.} To generate the test scenarios, we leverage the NSGA-II algorithm with two objectives. The first objective evaluates the deviation of the observed system's behaviour from its expected behaviour. %$for a particular test case. %The observed system behaviour is computed using an approximated system model, while the expected behaviour is specified in the requirements. 
The second objective is the test case diversity, calculated as a Jaccard distance with a reference test case.
To guide the first objective we are using a simplified system model rather than the full model. The full model is used to run the system in the simulation environment and can take substantial time to execute (several minutes for one scenario). The simplified system model is derived from the full model and can be used to get an approximation of the results obtained from the full model without running the simulation.\\
\textbf{Results.} We evaluate AmbieGen on three scenario generation case studies, namely a smart-thermostat, a robot obstacle avoidance system, and a vehicle lane-keeping assist system. For all the case studies, our approach outperforms the available baselines in fault revealing and several other metrics such as the diversity of the revealed faults and the proportion of valid test scenarios.\\
\textbf{Conclusions.} %\Foutse{first concludes on the suitability of nsga-ii for solving the problem before highlighting key lessons learned from the study!}
AmbieGen could find scenarios, revealing failures for all the three autonomous agents considered in our case studies. We compared three configurations of AmbieGen: based on a single objective genetic algorithm, multi-objective, and random search. Both single and multi objective configurations outperform the random search. Multi objective configuration can find the individuals of the same quality as the single objective, producing more unique test scenarios in the same time budget. 
%In order to produce diverse test scenarios, rather than a single scenario \Foutse{what do you mean by single solution? you mean solution based on a single objective?}, a second objective, accounting for diversity should be added. %allows producing test suites with fault revealing cases. %Comparing to the single-objective configuration
 %For the problems where one solution is enough 
 %\Foutse{what are these problems? how will people know in advanced that one solution is enough? please elaborate, otherwise it is vague!!! you said before that a second objective is important to ensure diversity!}% and there is a possibility to run the search multiple times, we suggest using the single-objective configuration of our approach as for some tasks it might converge faster to a higher fitness value. %for some
 %The approximated system \Foutse{what is the approximated system? you haven't defined it!!!} model can be used to reveal faults in the full system model \Foutse{what is the full system model? at this point in the paper none of those concepts are defined!!! you should first define concepts before using them!!!!}.
Our framework %is fine-tunable to different test scenario  generation problems and 
can be used to generate virtual environments of different types and complexity and reveal the system's faults early in the design stage. %AmbieGen allows to reduce the time for creating a search based environment generator for testing autonomous agents \Foutse{how do you know that it reduces...did you compare with a baseline?}.  %by researchers as a preliminary framework for testing the autonomous cyber-physical systems.
%\Foutse{Add a sentence about the potential benefits of your algorithm...explaining how it can be leverage both for further research and in practice!}
\end{abstract}

\begin{keywords}
cyber-physical systems,\\ test scenario generation, \\genetic algorithms, \\ virtual environments
\end{keywords}

\maketitle

%\vspace*{-1cm}

\section{Introduction{\label{sec:introduction}}}

One of the rapidly developing families of cyber-physical systems (CPS) are autonomous and vision based CPS. Examples include drone swarms, self driving cars, cave or underwater exploring robots.
%These systems are safety critical.
%The level of complexity of  such systems is high and so is the effort for their testing.
Typically, in the CPS development process the systems are validated and verified according to the V-model approach \cite{aerts2017model}. Prior to running the tests on a real system, the V-model includes model-in-the-loop and software-in-the-loop testing stages.
In these stages the simulations of the system are run in a virtual environment.
%is performed in the first place \cite{matinnejad2015search}.  \Foutse{to?...please elaborate and provide references!}  \Foutse{to? for? please elaborate!}. 
The goal is to model the real environment effect on CPS(s) and generate test scenarios violating some critical properties of CPS. However, during these simulations, engineers often lack tool support for generating the scenarios \cite{ben2016testing}. For particular applications there exist content generation techniques, like a Kruskal's algorithm for maze generation \cite{cormen2009introduction} or pre-configured scenarios, like the virtual worlds used in computer games \cite{sturtevant2012benchmarks}. However, they do not always provide the needed scenario complexity and  % or use the existing procedural content generation techniques, which are used to generate scenarios for the gaming industry and might not be effective enough to reveal the system faults. %One of the limitations of existing simulation environments, is that no guidance is provided to engineers 
%guidance \Foutse{do you mean tool support or guidance on scenarios? please clarify!!!} on which test scenarios to use  \Foutse{can you add a reference?}.  
oftentimes the scenarios have to be designed manually. 
Consider an autonomous robotic system, that should navigate to a goal destination in an environment with obstacles. The robots interact with the physical world via sensors and actuators in a feedback loop, avoiding the obstacles and searching the goal destination. Test scenario for such system includes virtual environment with obstacles , including moving and unexpected obstacles, changing terrain structure and environmental conditions. Manually designing all the possible scenarios in the virtual environment is a tedious task. %\Foutse{above you said that they lack guidance, here you seem to suggest that they lack automatic tool support instead...please check consistency!!!!!}. 

In this paper, we propose a search based approach, further referred to as ``AmbieGen`` to automatically generate test scenarios for cyber-physical systems. In the literature,  typically one objective, accounting for the scenario fault revealing power \cite{gambi2019automatically}, or two objectives, accounting for fault revealing power and diversity \cite{riccio2020model}, are used. To evaluate the contribution of adding the first and the second objectives we consider three configurations of AmbieGen: based on random search, single-objective genetic algorithm (AmbieGen SO) and multi-objective genetic algorithm NSGA-II (AmnbieGen MO). Preliminary results confirm that using the two objectives, maximizing both: scenario fault revealing power and diversity,   %The approach leverages the \Foutse{which one? any multi-objective algorithm?} NSGA-II multi-objective search algorithm with two objectives: first, maximizing scenarios fault revealing power and second, maximizing their diversity
allows to find more unique fault revealing scenarios given the same time budget. %The importance of these two dimensions was previously discussed in \cite{riccio2020model}. \Foutse{why are these two dimension important? please justify or at least add a reference to other works where their importance is discussed!!!}. 
To calculate the first objective we are using the simplified system model, derived from the full model of the system, as suggested by Menghi et al. \cite{menghi2020approximation}.The full model is used to execute the scenarios in the simulation environment and is usually computationally expensive. The simplified model allows to reduce the computational and time cost needed to produce the test scenarios as it does not require the simulator to run and provides the approximated outputs of the full model in a reduced amount of time.  %\Foutse{what is the simplified system model? you never defined it!!! You should define concepts before using them!!!}  \Foutse{what is this model???}.\Foutse{why? please conceptually explain the idea behind your approach before diving in low level details!!!}.
%\Foutse{The proposed approach leverages....(please provide some details about the conception of the approach!)}.
%and the system surrogate models to
%To improve the effectiveness of the initially \Foutse{which initially generated test cases? you never mentioned them before, please clarify!} generated test cases, we are using a multi objective search algorithm with two objectives: test difficulty and diversity. 

We evaluate our approach on three test generation case studies. In the first case study,\Dima{a smart-thermostat agent should follow the schedule with the expected precision. The goal of the testing approach is to generate the temperature schedule and a combination of environmental conditions violating the requirement. In the second case study, an autonomous robot should navigate to a goal location in an indoor environment with obstacles, without bumping into them. A test scenario should find such an indoor environment that makes the robot violate the requirements. In the third, an autonomous vehicle, with a lane-keeping assist system (LKAS), should follow a road lane of the given trajectory, without going out of its bounds. The testing approach should generate a virtual road that forces the car to go out of the road.}
%\Foutse{is this your solution to this problem? or it is the description of the problem? please describe the three problems consistently using similar structure and tense for sentences!!!} 
AmbieGen could reveal on average 9 failures in two hours for the autonomous robot model and 14 failures in two hours for the self-driving car model \Dima{executed in the simulators}.
Given the same evaluation budget, in all the case studies the multi-objective and single-objective configurations of AmbieGen produced fitter solutions with a large effect size, comparing to the random search. Multi-objective configuration allowed to produce more diverse scenarios with medium to large effect size comparing to single-objective, while finding solutions of the same or similar quality. %Considering the state of the art approach, Frenetic \cite{frenetic}, AmbieGen produced a similar number of fault revealing scenarios of 14 given a 2 hour time budget. 
%\Foutse{can you give some tangible evidences of this? some quantitative numbers? simply saying your results confirm the effectiveness is vague!!!}.

\textbf{This paper makes the following contributions:} 
\begin{enumerate}
    \item We design a search-based framework for generating customized environments for testing autonomous  %\Foutse{you included vision based before, please be consistent!!!} 
    CPS.
    \item We propose a novel technique for generating the virtual roads and robot navigation maps.
    \item Finally, we provide the code for replication of all of our experiments \cite{Ambiegen}.
\end{enumerate}%of this paper include: 
 %\Foutse{can you list the contributions as items!!! and please provide details for each of these contributions!!!}\Foutse{what do you mean by 'corresponding'???}  

Researchers and practitioners can leverage AmbieGen to automatically generate scenarios for autonomous CPS that will be further passed to the simulators to run tests on the full CPS models.  %\Foutse{please add a sentence to explain the potential benefits of your contributions!!!  Also use a consistent terminology to refer to your proposal...is it an algorithm or an approach or a techniques? pick one term and use it consistently throughout the paper!!!}

\textbf{The remainder of this paper is organized as follows.} Section \ref{sec:literature} discusses the related works in the domain of CPS testing. In Section \ref{sec:problem} we formalize the problem of scenario generation and provide the description of AmbieGen approach in Section \ref{sec:description}. Section \ref{sec:case} describes the test generation case studies used to evaluate our approach. In Section \ref{sec:eval} we formulate the research questions and our evaluation methodology. The same section reports our results and answers to research questions. Section \ref{sec:disc} discusses the results and the main challenges of this study. 
In Section \ref{sec:threat} we explain the threats to the validity of our results. Section \ref{sec:conc} concludes the paper and discusses some avenues for future works. %\Foutse{please provide an overview of the paper by briefly summarizing the contents of different sections. You can check the example of other papers on our lab website! } 
%A cyber-physical system consists of a collection of computing devices communicating with one another and interacting with the physical world via sensors and actuators in a feedback loop. 
%When designing and implementing autonomous systems, such as self-driving cars, flight controllers or robots,  a high level of quality assurance  of the system is critical. Before the 
%The complexity of such systems is increased due to the number of parameters they should monitor and control.
%To begin with, let's consider a wirelessly controlled robot. The robot monitors the environment and sends data to the operator, who takes decisions on the robot further movement.
%To test such system,  one should make sure that no combination of possible inputs from the operator would violate the robot safety requirements.

\section{Related literature {\label{sec:literature}}}
%The most common approaches for the CPS testing are  temporal logic falsification, Black-box testing, however, requires to execute the model under test for a large number of candidate test inputs. statistical
%model checking methods for Cyber-Physical Systems

%generating falsifying inputs for CPS, CPS test suite generation and IoT network protocol testing.
%CPS falsification testing.
%Monte Carlo sim-ulation (MCS), accelerated evaluation method (AE), worst-case scenario evaluation (WCSE)
Typically, the cyber-physical systems are developed using a model-based design approach \cite{aerts2017model}: after establishing the requirements of the system, model-in-the-loop testing is performed. In this step, models of the hardware part and the controller are created and tested in the simulation environment. 
One of the limitations of simulation platforms is that they do not provide clear guidance to engineers as to which test scenarios should be selected for simulation. Therefore, a number of approaches have been developed to generate the testing scenarios.

\textbf{General approaches for CPS testing}.
In the classical approach, the exhaustive exploration of the state-space of the model is performed \cite{broy2005model}. It uses the abstract model, created strictly according to the system requirements, to  generate the test cases for the model of the system under test (SUT). If the outputs of the SUT model and abstract model are different, the fault in the SUT is revealed.
As the system models get more complex, the search space becomes infeasible.
More recently, falsification based approaches have been proposed, verifying whether the model meets specific requirements specified in a temporal logic notation such as \Dima{timed computation tree logic (TCTL), linear temporal logic (LTL), metric temporal logic (MTL) or signal temporal logic (STL)}.
The \Dima{UPAAl SMC tool} performs the statistical model checking (SMC) of  of a given model, with the requirements specified using\Dima{TCTL notation}\cite{david2015uppaal}.
The core idea of SMC is to monitor some simulations of the system, and then compute the probability along with confidence
intervals that a specific requirement holds for the SUT.
A number of tools were developed that instead of calculating the probability that a system satisfies the property with a certain confidence, compute the worst expected system behaviour as a quantitative value, called robustness.
Examples of such tools are S-Taliro \cite{annpureddy2011s}, Breach \cite{donze2010breach} and ARIsTEO \cite{menghi2020approximation}. Differently from others, ARIsTEO propose to apply falsification testing to the surrogate, i.e.,  approximated model of the SUT, that closely mimics its behaviour but is significantly cheaper to execute. 
Arrieta et al., proposed a search based approach, that does not use the system model \cite{arrieta2017search}. They defined three cost effectiveness measures to guide search towards generating optimal test cases: requirements coverage, test
case similarity (effectiveness) and test execution time (cost).
%The common disadvantage of the mentioned approaches is that
%they search for fault revealing inputs, not relating the real world to the models under test. 
%When testing autonomous systems

\Dima{In the described works, authors consider falsification of the model of the system that takes as an input a set of signals $U = \{u_1, u_2, …, u_m\}$  and produces a set of signals $Y = \{y_1, y_2, …, y_n\}$ as the output.
In our work we focus on testing autonomous systems, for which the input signals are rather complex and might represent sensors and camera data, coming from different sources. Imagine a self-driving vehicle, using the lidar sensors and RGB camera to perceive the environment. Directly generating a valid combination of falsifying input signals (represented by lidar readings and RGB camera readings) would be rather challenging. Therefore, we focus on generating test cases that specify a virtual environment for an autonomous system, rather than the input signals. The input signals are generated in the virtual environment during the simulation, based on the actions taken by the autonomous agent.}A number of approaches have been proposed for generating virtual environments for testing the autonomous driving and robotic systems.%testing autonomous vehicle driving systems and robotic systems accounting for interaction with the environment.

\textbf{Vehicular driving system testing.}
Abdessalem et al. in \cite{ben2016testing} use a multi-objective search evolutionary algorithm  NSGA-2 to obtain test scenarios for Automated Emergency Braking (AEB) system.\Dima{They represent the test case as a tuple with sets of static parameters, such as precipitation, fogginess, roadshape and visibility range, as well as dynamic parameters, such as initial vehicle speed, initial pedestrians speed and location. The disadvantage of such encoding is the limitation in describing the static parameters. Such encoding does not allow to specify for example a complex road topology, with a number of turns and intersections. Also, we cannot account for the cases, where the static parameters need to be changed during the test case. For example, when the road material changes along the route, i.e., asphalt surface changes to pavement.}
 In order to generate challenging test cases, they use  fitness functions minimizing the distance to the pedestrian and time to collision, calculated while executing the test case.
To mitigate the computation cost of executing physics-based  simulations, the simplified models of ADAS trained on neural networks (NN) are used. Evidently, the NN has to be retrained each time model parameters are changed.

AsFault tool was proposed to generate the road configuration to test car Lane Keeping Assist System (LKAS) by Gambi et al., \cite{gambi2019automatically}. \Dima{This system should allow an autonomous car to always drive inside its lane. The test cases in this work correspond to a road topology that a car needs to follow.
AsFault represents roads as set of polylines, i.e., discrete sequences
of points. One polyline corresponds to one road segment. Roads are generated procedurally by stitching one road segment
to the next one. %By combining the polylines it generates a valid road network.
To evolve the roads, authors use a customized mutation operator, that randomly replaces a road segment and crossover operator that splits
two roads at a random point and recombines them, or recombines
random subsets of two roads. With such encoding combining the road polylines to produce a valid road may be challenging. Authors mention that if the search operator produces invalid road topology,  AsFault retries the application of the same search operator
with a probability of giving up which increases per failed attempt. Therefore, one disadvantage of such encoding is the difficulty in applying crossover and mutation operators. Another disadvantage is the limitations in encoding some important details of the test case such as the other vehicles, obstacles or pedestrian locations. Authors do not mention the possibility to encode such information with AsFault.}Finally, the selected fitness function maximizes the distance of ego-car from the center of the road lane. The real simulation model is used for calculating the fitness function value, which increases the computation cost.\Dima{ To improve the efficiency of the tool, authors filter out similar test cases by calculating Jaccard index between them. However, they do not maximize the diversity of the test cases explicitly.}

\Dima{Riccio and Tonella}\cite{RiccioTonella_FSE_2020}\Dima{ propose DeepJanus, a tool to explore the behavioural
space of a deep learning (DL) system to find pairs of inputs at its frontier: one input on which the DL system behaves as expected, and another similar input on which it misbehaves. In this work they consider two case studies: image classification (from MNIST database) and steering angle prediction for a vehicle LKAS system. We will further focus on the later. The scenario for the vehicle is represented as a road topology and it is encoded as a list of coordinates of the control points. These control points are further interpolated with Catmull-Rom cubic splines to produce a road topology. This encoding is limited in terms of increasing the complexity of the test case, such as specifying the location of other vehicles, obstacles or intersections. The tool is leveraging a two objective NSGA-II algorithm to generate effective test cases. Each individual of the algorithm  is encoded as a pair of road topologies. The first objective, $f_1$, aims to promote diversity between an individual and existing individuals as well as minimize the difference between the pair of road topologies constituting the individual. Authors define a problem specific distance metric used to evaluate the $f_1$. For the LKAS case study it is calculated as the Levenshtein distance between the roads. For image classification, it is calculated as an Euclidean distance between pixel matrices. The second fitness function, $f_2$, aims to minimize the distance of an individual to the behavioural frontier. In the case of the LKAS system case study it's calculated based on the maximum deviation from the lane center. If this deviation is higher than a certain threshold for one of the roads of the individual, the $f_2$ is assigned a -1 value, otherwise it's equal to the absolute value of the maximal deviation. We can see that the fitness functions are designed to evaluate a pair of road topologies, rather than a single road topology.}

\textbf{Autonomous robot testing approaches.}
Arnold  et al. \cite{arnold2013testing} designed a tool to produce navigation maps to test autonomous robot control algorithms. They use the Perlin noise to randomly generate the maps and then select the ones, that correspond to  a defined set of unwanted behaviours, such as stalling or colliding with a wall. No optimization algorithms are used to increase the number of fault-revealing scenarios.

Sotiropoulos et al. \cite{sotiropoulos2016virtual} generate navigation maps of different difficulty level  by changing the number and position of obstacles on the map. As in the previous work, the procedural content generation  with fixed parameters is used without optimization.

\Dima{Nguyen et al. \cite{nguyen2012evolutionary} propose a genetic algorithm based approach to design test scenarios for a cleaner robot. The robot goal is to collect all the rubbish given a
specific amount of time, without bumping into obstacles. The scenario represents an area with obstacles and objects to collect. It is encoded as a table with R x R cells. A cell containing an object is denoted
by 1, while a content–free cell is denoted by 0. The limitation of such encoding is that it is problem specific, i.e., it is suitable for encoding a map for the robot, however it is rather difficult to represent a road topology using such encoding. The fitness function is defined as closest distance to obstacles observed by the robot.}

\Dima{
Considering the existing approaches for environment generation, we can see that their implementation, including the evolutionary algorithm representation, has such limitations as problem specific or hardly customizable individual representation, difficulties in search operator implementation. In our work we propose a test generation approach, AmbieGen, with encoding that can be applied to test generation tasks for different autonomous agents, can provide a customizable scenario complexity level and allows to easily implement the search operators.
Furthermore, in our approach we are using two objectives, accounting for both fault revealing power $F_1$ and diversity of the test cases $F_2$ . Implementation of $F_1$ is problem specific and is defined as the difference between the expected and observed behaviour of the the agent. Contrary to existing works, we are explicitly promoting the diversity of the test cases as the second objective $F_2$ of the genetic algorithm using a universal distance metric, explained in Subsection \ref{subsec:objectives}. In our experiments we show that adding the $F_2$ allows to increase the diversity of the test cases in the resulting test suite, without reducing their fault revealing power. Overall, we surmise that our framework can be leveraged for implementing search based algorithms for test case generation for autonomous systems.}

%is limited fine-tuned to generate scenarios for a particular system. Redesigning the solution encoding and search operators for each new scenario generation task is tedious.
%Moreover, in some approaches, the evolutionary optimization is not applied.
%In our work we propose AmbieGen,  a search based test generation framework,  applicable to the generation of different types of virtual environments.
%with static and dynamic parameters, that can be fine-tuned for a particular task.
%We surmise that our framework can be used to design evolutionary algorithms for test environment generation. %by other researches and fine-tuned to their testing tasks.

 \section{Problem formulation{\label{sec:problem}}}
%\Foutse{can we expand on the formulation of the problem?}
%\Dima{Sure!}
A cyber-physical system is a reactive system, consisting of a computing device (or a collection of them) interacting with the environment via inputs and outputs \cite{alur2015principles}. \Dima{In a simulator, the autonomous CPS agent observes the virtual environment at discrete timestamps  $t_{i}$ using its local sensors, e.g., temperature, lidar sensors or RGBD-cameras, and measures its internal state $s_{t}$. At each timestep $t_{i}$ it receives a new observation $o_{t}$ $ \in  O$ and selects an action $a_{t}  \in A$ according to its control policy $\pi(s_{t}, o_{t}$). Here $A$ is a set of possible actions and $O$ is a set of all possible observations (the observation space), defined by the environment where the agent operates. Each test case represents an environment for the agent as well as the mission it needs to accomplish in it. We can formulate the problem as finding a test case, forcing the agent to take such actions $a_{t}$ that lead to violation of the established requirements and failure of the mission.}

For example, to test a self-driving vehicle we should consider such parameters as the road type and size, the location of other vehicles or pedestrians, the driving weather conditions, etc. Evidently, one of the most important requirements would be the collision avoidance.
Generating such scenarios manually is an extremely time consuming task: the engineers would need to list the precise positions and heading directions of the moving objects, specify different types of environmental conditions, etc. Therefore it is preferable to generate such scenarios automatically. To control their quality, optimization techniques should be used.
\subsection{Scenario representation}
In this subsection we formalize the definition of the test scenarios.\Dima{As it was mentioned earlier, one test case represents an environment in which the agent operates as well as the mission it needs to accomplish. We encode the test case as a set of parameters, needed to generate the defined environment in the simulator.} %For instance, to generate a road topology to test an 

\Dima{
First, we propose to divide the test case into at most $m$ parts. Each part can represent some aspect of the mission to accomplish or a part of the virtual environment in which the agent operates. We call these parts  the environmental elements $E_i$. We describe each environmental element with $n$ parameters, which we further refer to as attributes. }

\Dima{
As an example, let's consider a car lane keeping assist system (LKAS). It's main goal is to keep the vehicle within the road lane. One of the possible test cases to test this system can be a road topology, that the vehicle needs to follow. To design the test case, we suggest representing the road topology as a combination of road segments of different length and curvature. Here, each road segment would correspond to one environmental element and it's parameters such as length and curvature - the attributes of the environmental element.}

\Dima{Each individual (test case TC) is represented as a $n x m$ matrix where the cell $(i,j)$ contains values defining the TC.  In a nutshell, a TC has $n$ attributes $A_i$, and is composed of at most $m$ elements.  More precisely, cell $(i,j)$, alias $E_j$ (in line $A_i$), is the  value sampled from the attribute \textbf{$A_i$}. %needed at step j of the test case. 
Without loss of generality, an attribute  $A_i$, takes value into a set of possible realizations $\{a_0, a_1, ..., a_n\}$ or is defined as a  closed interval  $[A_{imin}, A_{imax}]$. We allow for the cells to be empty, if needed.}  % a TC is an ordered sequence of  consistent and meaningful  elements. It is the task of the fitness function  penalize inconsistent TC.   }

\Dima{Turning to the LKAS example, we can describe each road segment, i.e., environmental element, with such parameters as the length, the turning angle, the slope, the road material, etc. Each road segment will have a value for each of these parameters. By providing combinations of road segments with different parameter values, we generate various road topologies.
}

\begin{table}[h]
\begin{center}
\caption{Scenario $TC$ representation }
\label{tab:therm_repr}
\scalebox{0.95}{
\begin{tabular}{|c|c|c|c|c|c|} 
 \hline
&$E_1$ & $E_2$ & $E_3$ & ...& $E_m$\\
\hline
$A_1$&$A_{1e1}$ & $A_{1e2}$ & $A_{1e3}$ & ...&$A_{1em}$\\
\hline
$A_2$&$A_{2e1}$ & $A_{2e2}$ & $A_{2e3}$ & ...&$A_{2em}$\\
\hline
...&... & ... & ... & ...&...\\
\hline
$A_n$&$A_{ne1}$ & $A_{ne2}$ & $A_{ne3}$ & ...&$A_{nem}$\\
\hline
\end{tabular}}
\end{center}
\end{table}
Finally, the test scenarios can have restrictions $R$, which limit the scenario length $M$ or particular combinations of attributes.\Dima{In the LKAS example, for instance, we can limit the total road length or a certain combination of road segments, that create too sharp turns, etc.}

We provide examples of application of this representation to generate different test scenarios for autonomous agents in the Section \ref{sec:case}.

\Dima{Having obtained the matrix with attributes, i.e., the test scenario specification, we can generate the corresponding virtual environment in the simulator. According to the selected timestamp in the simulator settings, the agent will receive the inputs in the form of observations at each timestamp.  Observations can be represented by a temperature or lidar sensor readings, images from the RGB camera. After receiving each observation, the agent performs an action. Depending on the actions the agent takes it can fail or pass the test case.}

\Dima{In the LKAS example, suppose the car uses its RGB camera images to define its steering wheel angle. The standard camera frame rate to provide stable image for such applications is more than 30 Hz \cite{handa2012real}. Let's consider 30 Hz for this example. This means the car receives an input  in the form of the RGB image every 33,3 msec. For each received input, it provides a value of the steering angle to use. This value can be obtained, for instance,  by querying a neural network based model that classifies steering angle values based on the image of the road. }

Initial encoding of the scenario is the matrix $TC$, containing a high-level description of the environment. With such representation it's easy to implement the evolutionary search operators by simply\Dima{swapping the matrix columns between each other and randomly changing the cell values. By swapping the columns, we imply reassigning the attributes from one environmental element to another.}
To evaluate its fitness it first needs to be converted to the environment configuration for the approximated model ("TC to environment" module). %It should also be possible to convert it to the environment for the full model in the simulator. 
For example, the scenarios for an autonomous robot should be converted to a list of obstacle coordinates in a map. 
Then an approximated model is used to execute the scenarios and the fitness function is calculated based on the execution results ("Fitness function" module). Approximated model can be created either from real data or from the full model data. Another possibility is to \Dima{use a simplified system model, based on already implemented robotics algorithms,} such as those available at python robotics project \cite{robotics}. 
Overall, to customize AmbieGen, the developer needs to provide a list of attributes and their allowable ranges, an implementation of the "TC to environment" and "Fitness function" modules. 
The AmbieGen will integrate the modules and implement the initial population generation, crossover and mutation operators. 
Moreover, it's simple to control the level of complexity needed for the environment. By adding more attributes, the complexity can be increased. The limitation is the possibility of the simulator to interpret more complex environment configurations, such as the terrain type, the weather conditions, etc. 

\subsection{Search objectives definition}\label{subsec:objectives}
The main goal is to find scenarios producing system faults. At the same time, the scenarios should be diverse, uncovering different types of faults. From our experience, using only one objective results in producing many similar test cases in the last generation. Therefore we suggest adopting a multi-objective algorithm, where one of the objectives is accounting for the diversity of the test cases. The idea of adding a second objective for diversity is not new and was addressed in the novelty search works, such as \cite{mouret2011novelty}, and test scenario generation tools \cite{riccio2020model}.

To estimate\Dima{the first objective $F_1$,}the fault revealing power $\varphi$ of the test case, we compute the difference between the expected  $B(TC)$ and observed system behaviour $B_o(TC)$ after executing the test case:
\begin{equation}\label{f1}
   F_1 = \varphi(TC) = {\delta(B(TC), B_o(TC))}, \\
\end{equation}
%We introduce two goal functions to maximize both the fault revealing power and the diversity.
where $\delta$ is a developer defined function for computing the deviation between the expected and observed system behaviour and $TC$ is the test scenario specification.
The expected behaviour $B(TC)$ is typically defined in the system requirements or formulated by the developers e.g., "the car should not deviate from the lane center for more than 1 meter".
The observed behaviour corresponds to the model under test (MUT) outputs after scenario execution. However, the models of autonomous systems are rather complex and take long time to execute in the simulators, i.e., up to several minutes for one scenario. Moreover, executing the full models in the simulation environments requires additional system resources such as a GPU and high amount of RAM. Therefore we suggest estimating the observed behaviour $B_o(TC)$ %from test case execution 
using the approximated (surrogate) system models. Such models can be built based on the grey-box modelling approach \cite{romijn2008grey}, where model structure is chosen from system knowledge and parameters are selected to match sampled data. When little knowledge is available about the model, system identification techniques \cite{ljung1994modeling} can be used, where the modelled system is considered as a black box. 

%The process requires the following steps:
%\begin{enumerate}
%    \item Extract the data describing system behaviour in different modes.
%    \item Select a model structure.
%    \item Apply an estimation method to estimate values for the adjustable coefficients in the candidate model structure.
%    \item Evaluate the estimated model.
%\end{enumerate}

To estimate\Dima{the second fitness function $F_2$,} the variability $\upsilon$ of the test case, we compute the Jaccard distance between it and a reference test case:

\begin{equation}\label{jaccard}
    F_2 = \upsilon(TC, TC_{ref}) = 1 - \frac{TC \cap TC_{ref}}{TC \cup TC_{ref}}, 
\end{equation}
where $TC \: \cup\: TC_{ref}$ to the total number of\Dima{environmental elements}in both test cases and $TC\: \cap\: TC_{ref}$ to the number of inputs with similar or\Dima{same}attributes. %A similar metric was used ...

Finally, we define our search objectives:
\begin{gather*}\label{objectives}
    maximize: \varphi(TC), \\
    maximize: \upsilon(TC, TC_{ref}),\\
    subject \: to: C_1(TC) = \varphi(TC) - \alpha > 0,
\end{gather*}
Where $C_1$ is a constraint for the minimum value of the first objective,  $\alpha$ is the developer defined threshold to identify the test cases as having a risk of producing a failure.
This constraint is introduced to avoid producing test cases with low fault revealing power.

In our study we consider two configurations of our approach: AmbieGen MO, described above and AmbieGen SO based on a single objective genetic algorithm (GA) with $F_1$ as a fitness function.
 \section{Proposed approach description{\label{sec:description}}}
To perform the search we are using evolutionary search algorithms NSGA-II and GA \cite{coello2007evolutionary}, which have proven to be effective at similar tasks \cite{abdessalem2018testing, arrieta2017search}. Below, we present the GA and NSGA-II configurations used in AmbieGen. %In this subsection We present the algorithm pseudocode and its description.

%\subsection{Genetic algorithm configuration}
We implemented the AmbieGen MO and AmbieGen SO using a python Pymoo framework \cite{pymoo}. The framework provides the possibility to define custom solution representations, crossover and mutation operators. 
%\Foutse{we may not need to use subsections here since they are quite small....}

\textbf{Solution representation.}
Each individual in the population corresponds to a test case. Individuals can have a variable number of genes, i.e., environment elements depending on the application.
Internally, we represent the individual, i.e.,  the test case, shown in table \ref{tab:therm_repr}, as a dictionary, as shown below:
\begin{lstlisting}[language=Python]
{"E1": { "A0:" A0, "A1": A1, ..., 
"An": An }, "Ei": {...} }, 
\end{lstlisting}
where $Ei$, corresponds to the element of the environment that is described and \Dima{$An$ to the value of the attribute from the defined attribute set $A$.} %$M$ is in allowable range of values $M \in [V_{njMIN}, V_{nMAX}]$.

\begin{figure}[h]
\includegraphics[scale=0.85]{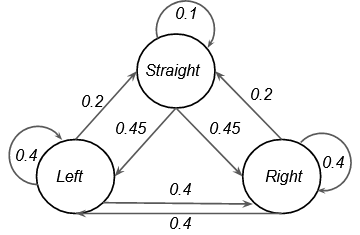}
\centering
\caption{Markov chain used to generate the intiial test cases for the LKAS case study}
\centering
\label{fig:markov}
\end{figure}

\textbf{Initial test case generation.}
The search begins by generating the initial test cases.
One of the options is to \Dima{assign arbitrary values to environmental attributes $A_0$, ..., $A_n$ of each element $E_i$ from their allowable value ranges  $[A_{imin}, A_{imax}]$ or realisation sets $\{a_0, a_1, ..., a_n\}$.} When some distribution of attribute values is known to  produce better test cases, from both semantical and fault-revealing point of view, we suggest using the Markov chain to assign the values \Dima{of certain attributes}. For example, when generating the road segments, the road consisting of only the straight segments is very unlikely to produce faults. Such cases can be avoided by assigning values with the Markov chain.\Dima{To build a Markov chain we need to define the state space as well as the probability table of switching between these states. As the state space we can use the possible values of one of the attributes. In the case of the LKAS it can be the three road segment types: going straight, turning left and turning right. The state switching probability table can be inferred from the domain knowledge. In the case of the LKAS system, we want to lower the probability of getting a sequence of only straight road segments. An example of a Markov chain used to generate the initial test cases for the LKAS  is shown in Fig.\ref{fig:markov}. }

\textbf{Fitness evaluation.}
We use two fitness functions to evaluate each individual: $F_1$, corresponding to the function in (\ref{f1}) and $F_2$ corresponding to (\ref{jaccard}).
 $F_1$ is calculated after executing the test case with a surrogate model $M$ of the system. This function is problem specific and should be proportional to the unwanted behaviour of the system. For example, for evaluation of a self-driving car test case we can compute the maximum deviation from the road lane center, where bigger deviation is likely to produce more faults.
 
 In our implementation we compute $F_2$ as the Jaccard distance between the individual and its parent, which acts as a reference test scenario. The intuition is to promote the modifications done to the test scenarios. However, a different reference test scenario can be used, such as the closest individual from the Pareto optimal solutions.

 As the Pymoo framework minimizes the fitness functions, in our implementation we multiply $F_1$ and $F_2$ actual values by (-1).
 
 \textbf{Mating selection.}
 To select the individuals for crossover and mutation the binary tournament selection is used, which is implemented by default in Pymoo. $N$ individuals are selected, producing $N$ new individuals after crossover and mutations.
 
\begin{figure}[h]
\includegraphics[scale=0.65]{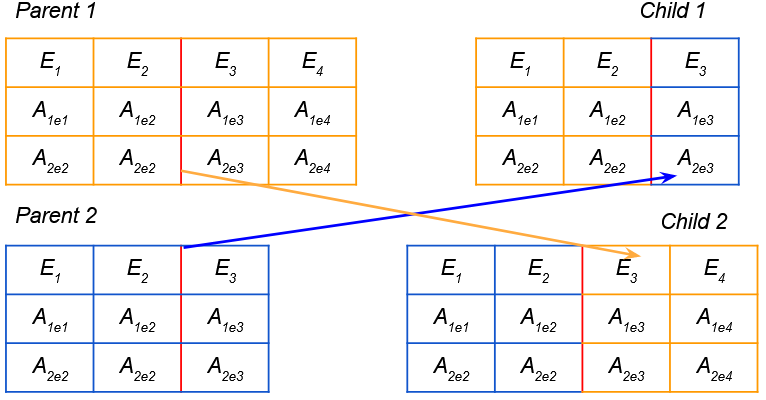}
\centering
\caption{Crossover operator for two test cases with 5 and 6 environmental elements with the crossover point at the third element}
\centering
\label{fig:cross}
\vspace{-5mm}
\end{figure}

\textbf{Crossover operator.}
We are using a one point crossover operator, which is one of the commonly used operators for variable-length solution representation. \Dima{This operator creates two new test cases by exchanging information between two existing test cases $TC_1$ (parent 1) and $TC_2$ (parent 2), with corresponding lengths  of $m_1$ and $m_2$. Let's suppose that $m_1$ is smaller than $m_2$. It is performed in two steps. First we randomly select the crossover point $k$ with the index from 1 to $m_1 - 1$.  Then the elements of the $TC_1$ with the indices from $k$ to $m_1$ and elements of $TC_2$ with indices from $k$ to $m_2$ are swapped. An illustration of the crossover operation between two individuals is shown in Fig. \ref{fig:cross}. Individual $TC_1$ length is 4 elements and individual $TC_2$ a size of 3. Both individuals have two types of attributes $A_1$ and $A_2$. $A_{iej}$ is a value of the attribute $A_i$ corresponding to the environmental element $E_j$. The crossover point is chosen to be 2 and is shown as a red line.
} 

\textbf{Mutation operators.}
We define three mutation operators:
\begin{itemize}
    \item \textit{exchange operator}: \Dima{the attributes of} two randomly selected environmental elements of a chromosome are  exchanged the positions;
    \item \textit{change of variable operator}: an environmental element $E_i$ in a chromosome is randomly selected, then for one of the attributes $A_n$ the value is changed according to its type and maximum as well as minimum values.
     \item \textit{scramble operator}: \Dima{attributes of a}number of environmental elements $E_i$ in a chromosome are selected, then their positions in the chromosome are randomly exchanged.
\end{itemize}
\textbf{Individual insertion.}
To insert the individuals the  mu+lambda approach is employed \cite{beyer2002evolution}. The idea is to merge the population and offsprings together, and then from the merged set, select the best possible non-dominated solutions of the population size. 
 \section{Test scenario generation case studies{\label{sec:case}}}
In this section we demonstrate how AmbieGen can be applied to three different types of environment. We consider the following test generation case studies:  a smart-thermostat, robot obstacle avoidance system, and vehicle lane-keeping assist system (LKAS). In every case study the autonomous agent controller has a different level of complexity: simple proportional–integral–derivative (PID) controller for the thermostat, a robot controller based on the nearness diagram navigation approach \cite{nd}, and a deep neural network based controller for the vehicle. We evaluate AmbieGen by comparing the results obtained with random search for all the three problems. For the LKAS system case study we also compare our results with state-of-the art approach, presented at SBST2021 tool competition\footnote{\url{https://sbst21.github.io/tools/}}.

 \subsection{Wireless thermostat case study{\label{sec:case1}}}

Nowadays, home automation becomes more and more popular.  Automatic temperature control systems are one of the most commonly used. Such systems consist of a controller, temperature sensor and a heating element. The controller goal is to keep the room temperature according to the programmed schedule. The simplest solution is to send "ON" and "OFF" commands to the heater, when the temperature needs to be increased or decreased. More sophisticated thermostats implement PID controllers to achieve smoother operation.  Testing the controllers in the simulators for different temperature schedules is necessary in order to ensure their precision and reveal the possible limitations.
In this study, the test generation goal is to create scenarios accounting for the scheduled temperature as well as environmental conditions.

\subsubsection{System under test description}
In our case study we consider a simple real-world wireless thermostat system. It consists of one room with a heater, sensor, and controller and is part of a larger system, described in more details in \cite{zid}. The room dimensions are approximately 2.5 m × 4 m and the height is about 2.6 m. The heating element is a Steelpro 1.5 Kw electronic convector \footnote{\url{2https://www.stelpro.com}}, which is controlled via a wireless Z-wave protocol based switch. The temperature is measured by a Aeotec MultiSensor 6 device \footnote{\url{https://aeotec.com/z-wave-sensor}}, placed at about 2.2 m from the floor. The controller is a Raspberry Pi 3B running Z-wave.me with a RaZberry 4 \footnote{{\url{https://z-wave.me/products/razberry}}} daughter card acting as Z-wave network controller. The Raspberry Pi has a user defined schedule of temperature levels; it reads the thermometer measured values and if needed (according to the schedule and required temperature) it switches on (off) the heating. 
 
\begin{figure*}[ht]
\begin{subfigure}{.497\textwidth}
  %\centering
  % include first image
  \includegraphics[scale=0.58]{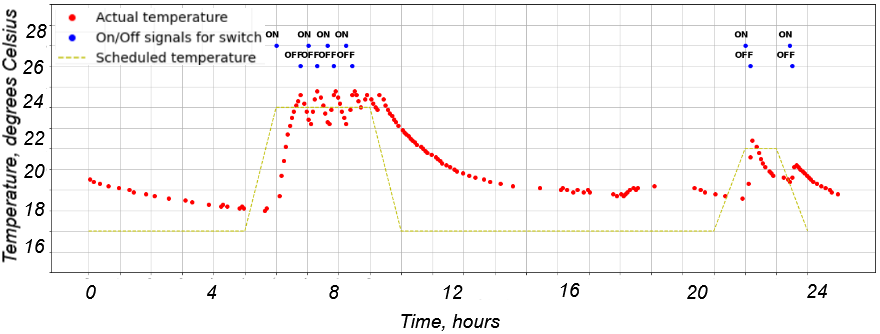}  
  \caption{Slower decreasing temperature pattern}
\end{subfigure}
\begin{subfigure}{.497\textwidth}
  %\centering
  % include third image
  \includegraphics[scale=0.58]{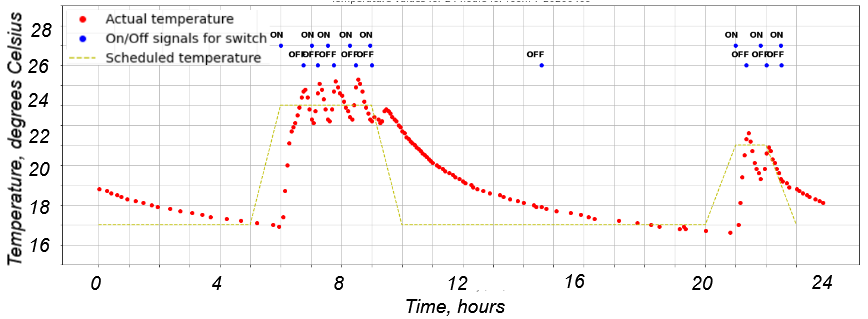}  
  \caption{Faster decreasing temperature pattern}
\end{subfigure}
\caption{Different temperature dynamics patterns}
\label{fig:fig_compare}
\end{figure*}

The data for the room temperature was collected for the period from December 2019 to May 2020. From the data, we could observe 7 patterns of temperature dynamics after ON/OFF commands of the thermostat. As an example, in the Fig. \ref{fig:fig_compare} you can see that the temperature decreases with different rates, which depends on such environmental factors as indoor and outdoor temperature, humidity, etc. We represent different temperature dynamics patterns, accounting for the different environmental conditions,  with different thermostat models. 

%The temperature in the room is set by the user defined schedule. In addition,
%the environmental conditions affect the system behaviour: the
%commands sent from the controller to the thermostat can be
%delayed or even lost due to the network overload, the temperature
%decrease/increase speed can vary depending on the time
%of the day, room humidity, etc. 

%Given a user defined schedule,
%the goal is to find the  scenarios when the system is not able to follow the schedule.
%\subsubsection{Case study motivation and context}

\Dima{The thermostat should be able to keep the scheduled temperature with the precision of 1 degree Celcius, under different environmental conditions in the room.
The goal of the search is to find the schedule and the corresponding thermostat operating mode, to falsify this requirement.}

\subsubsection{Problem representation}
For this problem we define three high-level input attributes: $A_1$ the goal temperature value, $A_2$ the duration of this temperature and the thermostat operation mode $A_3$.\Dima{Each test case contains $m$ environmental elements $E$. For each of them we need to specify the value of each of the attributes $A_1, A_2, A_3$.} The allowable ranges for the attributes are shown in Table \ref{tab:tab2}.
 
\Dima{An example of encoding of the individual representing the temperature schedule illustrated in Fig. \ref{fig:fig_compare} (yellow line) is shown in Table \ref{tab:individ1}. Each environmental element represents a part of the schedule, i.e, part of agent's ``mission'', as well as the operating mode.}
\begin{table}[h]
\begin{center}
\caption{Attribute types for thermostat problem}
\label{tab:tab2}
\scalebox{0.95}{
\begin{tabular}{|c|c|c|} 
\hline 
$A_1, temperature, T^{\circ}$ & $A_2, duration, min$ & $A_3,  operation \;  mode$\\
\hline
[16, 17,...,25], & [60, 75,..., 240] & [1, 2,..., 7 ]\\
\hline

\end{tabular}}
\end{center}
\end{table}
\begin{table}[h]
\begin{center}
\caption{Example of individual representation for the first case study}
\label{tab:individ1}
\scalebox{0.85}{
\begin{tabular}{|c|c|c| c |c |c |} 
%\hline
%\textbf{Original value}& \textbf{Encoded value} &  \textbf{Original value} & \textbf{Encoded value}\\ 
\hline 
 & $E_1$ & $E_2$ & $E_3$ & $E_4$ & $E_5$ \\
\hline
$A_1, \; goal \; temperature$ & 17 & 23 & 17 & 21 & 17\\
\hline
$A_2, \; duration$ &  5 & 4 & 12 & 2 & 1 \\
\hline
$A_3, \; mode$ & 1 & 3 & 2 & 4 & 1 \\
\hline
\end{tabular}}
\end{center}
\end{table}

We define two restrictions $R_1$ and $R_2$ for the test scenarios.
For $R_1$,  the duration of the schedule $T$ cannot exceed 24 hours:
\begin{equation} \label{req1_therm}
  R_1: \sum_{i=1}^{m} A_{2ei} < T, \; T=24\\
\end{equation}
\Dima{This equation represents the sum of the values of attribute $A_2$ (duration of the set temperature) of each of the $m$ environmental elements.}

For $R_2$, the temperature cannot change too sharply, i.e., more than 5 degrees between two adjacent\Dima{environmental elements}$E_i$ and $E_{i+1}$:
\begin{equation}\label{req2_therm}
   R_2: \mid A_{1e_i} - A_{1e_{i+1}}\mid < 5, i \in [1, m-1],  \\ 
\end{equation}
where $m$ is the number of\Dima{environmental elements}in the test case.
\subsubsection{Fitness function definition}

To calculate one of the fitness functions we need to create a simplified model of the system. 
To this end, we extracted the data  from the experimental measurements and selected the series of data points, corresponding to behaviour of the thermostat after "switch on" and "switch off" commands in different thermostat operation modes.
The next challenge is to select the model structure. In our case it is possible to build a first-principles model, as the heating and cooling of a closed space is guided by physical laws, such as Newton Law of cooling \cite{winterton1999newton}.  The law has an exponential nature, therefore our model structure is based on increasing and decreasing exponential function.

\begin{figure}[ht]
\includegraphics[scale=0.58]{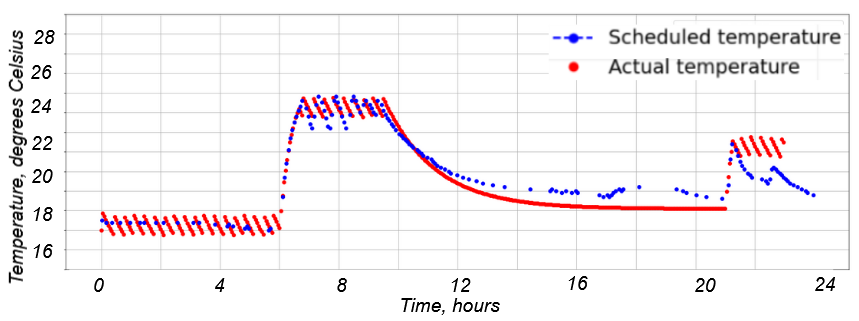}
\centering
\caption{Model (red points) fitting the experimental data (blue points)}
\centering
\label{fig:model_fit}
%\vspace{-5mm}
\end{figure}

We propose the following time-discreet model structure for the $M_1$ ("on") mode:
\begin{equation}
    Y = k_{on1}*(1 - e^{-k_{on2}*t_i}) + T_0
\end{equation}
and for the $M_2$ ("off") mode:
\begin{equation}
    Y = k_{off1}*(e^{-k_{off2}*t_i}) + T_0 - k_{off1}
\end{equation}
Here $k_{on1}$, $k_{on2}$, $k_{off1}$, $k_{off2}$ are the unique coefficients defining the model behaviour in a particular environment. $T_0$ - is the starting temperature and $t_i$ - the discreet time step value, $Y$ - the output temperature.
We keep the coefficients in a table, such as Table \ref{tab:tab1}, where coefficients for the three models are shown. As an example, in Fig. \ref{fig:model_fit} you can see how the model 1 with the coefficients from the table, fits the data from real measurements. One model includes two equations describing behaviour in "on" and "off" modes. In total, we identified 7 models having different coefficients in the equations, corresponding to the thermostat operating in different environmental conditions.
\begin{table}[h]
\begin{center}
\caption{Model coefficients}
\label{tab:tab1}
\scalebox{0.85}{
\begin{tabular}{|c|c|c| c | c|} 
%\hline
%\textbf{Original value}& \textbf{Encoded value} &  \textbf{Original value} & \textbf{Encoded value}\\ 
\hline 
Model & $k_{on1}$ & $k_{on2}$ & $k_{off1}$ & $k_{off2}$\\
\hline
1 & 7.7 & 0.11887928 & 5.6 & 0.02929884\\
\hline
2 & 7.9 & 0.11180434 & 5.2 & 0.04803319\\
\hline
3 & 6 & 0.14704908 & 4.8 & 0.1203876\\
\hline
\end{tabular}}
\end{center}
\end{table}

%Evidently, due to varying environmental conditions, i.e the opened door, higher or lower humidity, heat transfer from outside, the coefficients in the selected model structure had to be adjusted to better fit the original data. Creating one complex model, with high number of inputs, considering the environmental conditions, would make the execution of the model computationally expensive.

To obtain the coefficients, we fit the experimental data by a curve with minimal deviation. We used python SciPy library, namely $curve\_fit$ function from $Optimize$ class, which  is based on the non-linear least squares method \cite{scipy}.
The average root mean square error between original and approximated data did not exceed 0.5 degrees Celsius.

\Dima{Next, we specify the requirement for our system and introduce the fitness function based on this requirement: the root-mean square error between the scheduled temperature and the temperature set with the thermostat should not exceed 1 degree C$^{\circ}$.}

To calculate the first fitness function, $F_{1therm}$, we execute the test scenario $TC$ using the simplified model. We obtain the output values of the room temperature set by the thermostat $Y$ and calculate the root-mean square error between $Y$ and the temperature values defined in the schedule $S$:
\begin{equation}
    F_{1therm} =  \sqrt{\sum_{i=1}^{n}\frac{(Y_i - S_i)^2}{n} }, 
\end{equation}
where $n$ is the number of datapoints in the output.
For the test cases that do not satisfy the \Dima{restrictions} (\ref{req1_therm}) and (\ref{req2_therm}) we set the $ F_{1therm}$ to 0.

%\subsubsection{Scenario creation}

We calculate the second fitness function, $F_{2therm}$ according to (\ref{jaccard}).
In order to prevent obtaining the test cases with low fault revealing power, we also add a search constraint $C_{therm}$:
\begin{equation}
    C_{therm}: |F_{1therm}| - 1.5 > 0  , 
\end{equation}
%An example of the generated scenario is shown in the Fig.\ref{fig:them_tc}. We can see that due to the environmental conditions, there are periods in the day, when the thermostat is not able to follow the schedule with the required precision.

\subsubsection{Genetic algorithm configuration.}
We used the following GA (AmbieGen SO) and NSGA-II (AmbieGen MO) configurations for the smart thermostat problem: population size: 250, number of generations: 200,mutation rate: 0.4, crossover rate: 1, algorithm type: generational, number of evaluations: 50 000.

We are using a high mutation rate, as from our experience, it allowed to converge to better solutions faster. In our implementation a $\mu$+$\lambda$ insertion approach is used, where only the best individuals from previous generation and offsprings are inserted to the next generation. %Therefore, high mutation rate only broadens the search space as the individuals with lower fitness after mutation are discarded. 
In the generational GA the number of offsprings inserted in the population is equal to the population size. We limit the number of evaluations to 50000, as typically after this number was enough for the algorithm to converge. The average time to run 50000 evaluations was 136.691 sec for GA and 123.665 sec for NSGA-II. 

\subsubsection{Scenario generation}
\begin{figure}[ht]
\begin{subfigure}{.5\textwidth}
  \centering
  % include first image
  \includegraphics[scale=0.58]{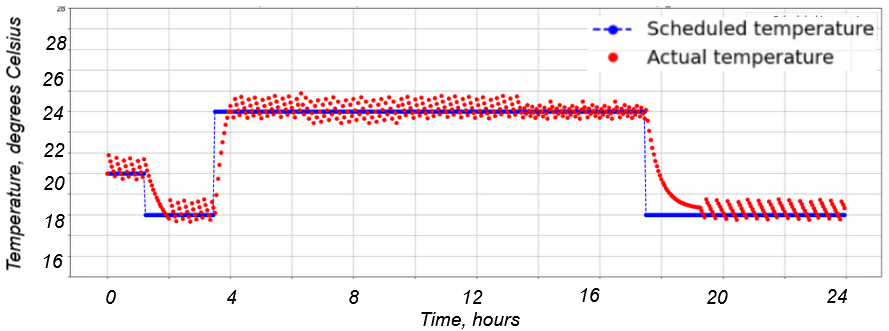}  
  \caption{Randomly produced scenario}
  \label{fig:tc-first}
\end{subfigure}
\begin{subfigure}{.5\textwidth}
  \centering
  % include third image
  \includegraphics[scale=0.58]{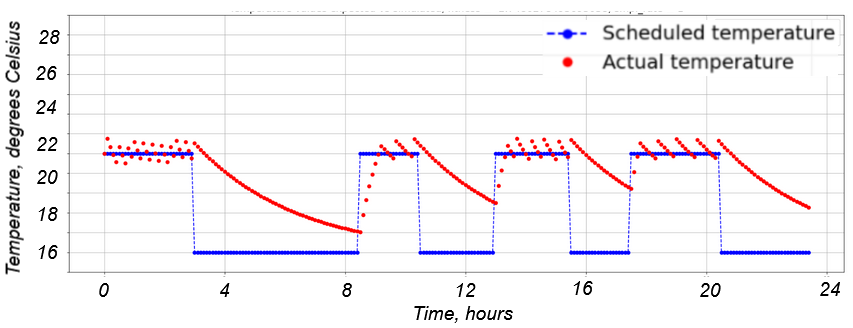}  
  \caption{Search produced scenario}
  \label{fig:tc-second}
\end{subfigure}

\caption{Examples of scenarios for the smart thermostat}
\label{fig:fig_tc_therm}
\end{figure}

Finally, we discuss an example of the produced scenarios.
In Fig.\ref{fig:tc-first} you can see a scenario with a low fitness value of 0.76 degrees, indicating that the temperature deviates from the schedule 0.76 degrees on average. On the contrary, in  Fig.\ref{fig:tc-second} the scenario produced by AmbieGen search has a higher fitness of 2.4 degrees. Clearly, this scenario is more likely to be unacceptable to the user, comparing to the first one.

 \subsection{Autonomous navigating robot  case study {\label{sec:case3}}}

%In this problem, the goal is to generate the test scenarios i.e. navigation environment for a simulated mobile robot.
The autonomous robotic systems are used in many domains: from everyday tasks such as room cleaning to critical missions such as navigation to harsh environments.  
For every application, we need to have a high confidence that their behaviour will be safe. Running the simulations of the system in various virtual environments can uncover the possible failures of the robot in the early design stage.

In this case study we consider an autonomous mobile robot, navigating in a space with obstacles. The robot has to reach the goal location, relying only on its range sensors and the planning algorithm. The goal is to generate the environment,  i.e., a room with obstacles that forces the robot to fail. Similar test generation problems were addressed by \cite{sotiropoulos2016virtual}, \cite{arnold2013testing}.
In \cite{arnold2013testing} the navigation maps are created using the procedural content generation technique. Then robots are assigned a randomized route to follow. The test scenario is an environment populated with robots, obstructions, and mission allocations. 
Sotiropoulos et  al. characterize a map by its size, percentage of obstruction (due to objects), and its degree of smoothness (resulting from the ground local deformations). The robot is given a navigation  mission, defined by a starting position and a target arrival position, situated in the map boundaries. 
Both approaches only consider the random generation.

\subsubsection{System under test description}
\begin{figure}[h!]
\begin{subfigure}{.23\textwidth}
  \centering
  % include third image
  \includegraphics[scale=0.3]{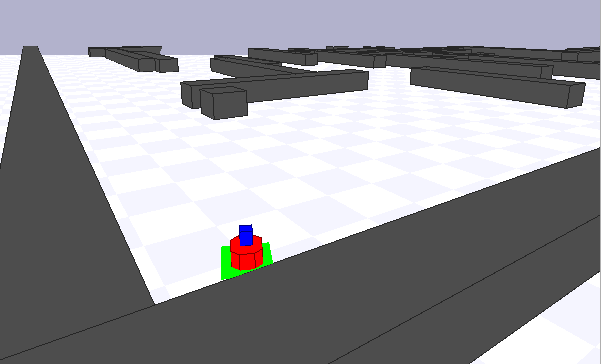}  
  \caption{A perspective view in the simulation environment}
\end{subfigure}
\begin{subfigure}{.23\textwidth}
  \centering
  % include fourth image
  \includegraphics[scale=0.33]{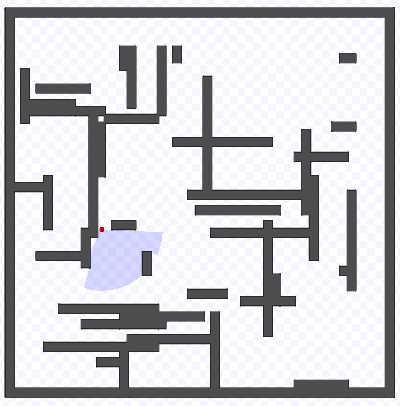}  
  \caption{A top view in the simulation environment}
\end{subfigure}
\caption{Player/Stage simulation environment for autonomous robots}
\label{fig:player}
\end{figure}

We ran the simulations in the Player/Stage simulation environment (see Fig. \ref{fig:player}), which is one of the most commonly used in the robotics field \cite{kranz2006player}. We also considered using such simulators as Gazebo\footnote{\url{http://gazebosim.org/}}, MORSE\footnote{\url{https://github.com/morse-simulator/morse}}, and Argos\footnote{{\url{https://www.argos-sim.info/}}}. One of the advantages of Player/Stage for our study was the possibility to load the automatically generated environment configuration files as well as the big number of implemented models and controllers. For Gazebo and MORSE the environments have to be manually created in a dedicated 3D design tool. For Argos, the maps can be generated automatically, however the number of implementation examples is limited. One of them, which includes a planning algorithm implementation, is dedicated to robot swarms, which we plan to explore more in the future \cite{varadharajan2020swarm}.

For simulations we used a Pioneer 3-AT mobile robot \footnote{\url{https://www.generationrobots.com/media/Pioneer3AT-P3AT-RevA-datasheet.pdf}} model provided by the Player/Stage simulator. The robot is equipped  with  a  SICK  LMS200  laser with the sensing  range  of  10 meters, it has  four wheels and is  capable  of  speeds  of  up  to  0.8m/s.
One of the planning algorithms provided by the simulator is 
using the nearness diagram (ND) navigation method. This is a reactive navigation method, where the motion commands are computed based on the robot sensor data. The method computes the optimal motion command to avoid collisions while moving the robot toward a given goal location. Before the robot mission starts, it runs the A* planning algorithm to obtain the route towards the goal. Then it uses its ND algorithm for navigation. To increase the challenge for the robot we applied the Ramer–Douglas–Peucker algorithm \footnote{\url{https://rdp.readthedocs.io/en/latest/}} to reduce the number of waypoints in the created route.

\Dima{Given the goal location the robot should navigate to it without bumping into obstacles. The testing approach objective is to find environments, when the navigation algorithm fails and the robot does not reach the goal or the robot hits an obstacle during the navigation.} The scenario is represented as a bitmap, where the location of obstacles is specified, as well as by a set of waypoints for the robot to follow. Failures are detected by a daemon script that continuously monitors the simulation environment.

%The Ramer–Douglas–Peucker1 algorithm is used to reduce the number of waypoints,  both  to  reduce  storage  space  for  the  situation  and  to  force  robots  to  do  some online global path-planning to avoid obstacles, thus increasing the challenge for their movement algorithms

%\Foutse{can we elaborate a bit more on the context of each case studies?}

\subsubsection{Problem representation}
In this scenario generation case study the environment is represented by a map with obstacles. We define the map size to be 50 x 50 m. Each environment part $E_i$ corresponds to a space of the size 1 x 50 m.\Dima{There is one obstacle in each environmental element.}In total there are 50  elements $E_i$,\Dima{and 50 obstacles in each test case.}The scenario matrix size $M$ is fixed and is equal to 50.
We define three attributes describing the environment: $A_1$, the type of the obstacle, $A_2$ position of the obstacle and $A_3$ the size of the obstacle. This gives the size $N$ of the scenario matrix of 3. 
The values for the attributes are specified in Table 
\ref{tab:rob_param}. We use two types of obstacles  - vertical and horizontal walls. The size is the total obstacle length in meters. The position is the obstacle center location in the element $E_i$. 

\begin{figure}[h]
\includegraphics[scale=0.32]{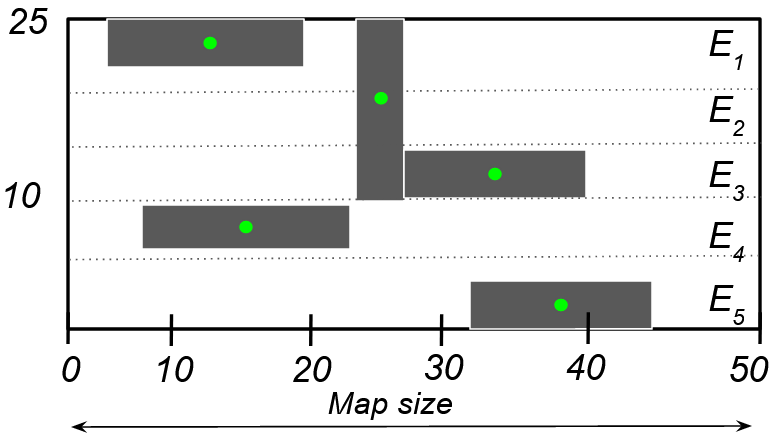}
\centering
\caption{An example of the test case represented by an individual in table \ref{tab:individ2}}
\centering
\label{fig:individ2}
%\vspace{-5mm}
\end{figure}

\Dima{An example of individual encoding that represents a map with obstacles in Fig. \ref{fig:individ2} is shown in Table \ref{tab:individ2}. Here the map size is 50 m x 25 m and each environmental element represents a part of the map of the size 50 m x 5 m. For each environmental element the location of the center of the obstacle (marked with a green circle) is specified. Horizontal obstacle is encoded with the value of 0 and the vertical obstacle with the value of 1.}

We define two restrictions. First, $R_1$: only one obstacle per element $E_i$. Second, $R_2$: the obstacles cannot cover completely or intersect with the initial and target robot location points.

\begin{table}[ht]
\begin{center}
\caption{Attribute types for autonomous robot problem}
\label{tab:rob_param}
\scalebox{0.95}{
\begin{tabular}{|c|c|c|} 
\hline 
$A_1, \; obstacle \; type$ & $A_2,\; obstacle\; size$ & $A_3, \; obstacle \; position$\\
\hline
[horizontal, vertical] & [5,6, ..., 15] & [1, 2, ..., 50 ]\\
\hline

\end{tabular}}
\end{center}
\end{table}

\begin{table}[h]
\begin{center}
\caption{Example of individual representation for the second case study}
\label{tab:individ2}
\scalebox{0.85}{
\begin{tabular}{|c|c|c|c|c|c|} 
%\hline
%\textbf{Original value}& \textbf{Encoded value} &  \textbf{Original value} & \textbf{Encoded value}\\ 
\hline 
 & $E_1$ & $E_2$ & $E_3$ & $E_4$ & $E_5$\\
\hline
$A_1, \; obstacle \; type$ & 0 & 1 & 0 & 0 & 0\\
\hline
$A_2, \; obstacle \; size $ &  15 & 10 & 10 & 15 & 10 \\
\hline
$A_3, \; obstacle \; position $ & 12 & 25 & 35 & 15 & 38 \\
\hline
\end{tabular}}
\end{center}
\end{table}

\subsubsection{Fitness function definition}

The intelligent robotic systems are typically equipped with a planning algorithm that builds a path to the goal location as the robot moves through the environment. The trajectory is adjusted as the new obstacles are discovered by the robot.

In the simplified case, the robot knows about the location of all the obstacles in advance. Therefore, as the robot approximated model we are using the Python robotics implementation of A* planning algorithm \cite{robotics}, which creates the route given the map, start and destination location. We have selected the A* because it is a deterministic algorithm and always finds a route, if it exists. The disadvantage is that the computations take longer time, than for non-deterministic planning algorithms such as RRT*.

\Dima{The requirement for our system is that the robot should navigate from the start to the goal location, without bumping into an obstacle. To falsify this requirement, intuitively, the test case should force the robot to follow a complex path to the goal location.}

The first fitness function, $F_{1robot}$, maximizes the distance the robot would have to travel to find the goal.\Dima{Evidently, travelling a longer distance, the robot takes a more complex path to the goal location, that involves a higher number of turns.}
For the test cases that do not meet the\Dima{restrictions}$R_1$ and $R_2$, $F_{1robot}$ is set to 0. 
 The second fitness function is calculated according to (\ref{jaccard}).

%\subsubsection{Scenario creation}
\subsubsection{Genetic algorithm configuration}
We used the following GA (AmbieGen SO) and NSGA-II (AmbieGen MO) configurations:
population size: 100, 
number of generations: 400,
mutation rate: 0.4,
crossover rate: 1,
algorithm type: steady state with 50 offsprings,
number of evaluations: 20,000.

For this problem we used a smaller number of offsprings to run more generations for the same time budget. The A* algorithm implementation was computationally expensive to execute.
The average time to run 20,000 evaluations was 2,727.2 sec for AmbieGen SO and 2,394.9 sec for AmbieGen MO.

\subsubsection{Scenario generation}
\begin{figure}[ht]
\begin{subfigure}{.23\textwidth}
  \centering
  % include third image
  \includegraphics[scale=0.28]{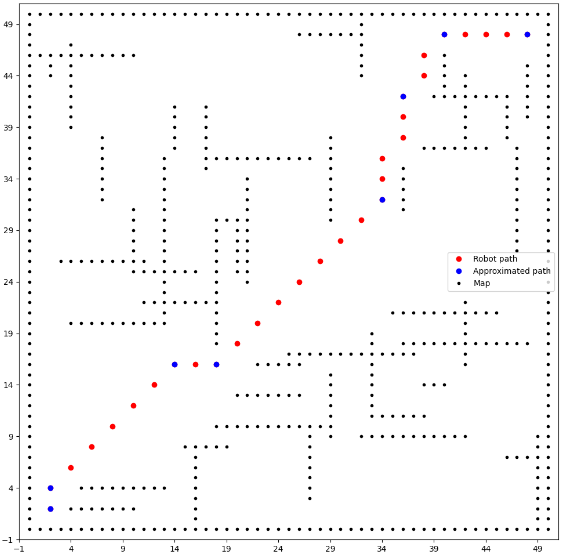}  
  \caption{Random scenario}
  \label{fig:sub-room1}
\end{subfigure}
\begin{subfigure}{.23\textwidth}
  \centering
  % include fourth image
  \includegraphics[scale=0.28]{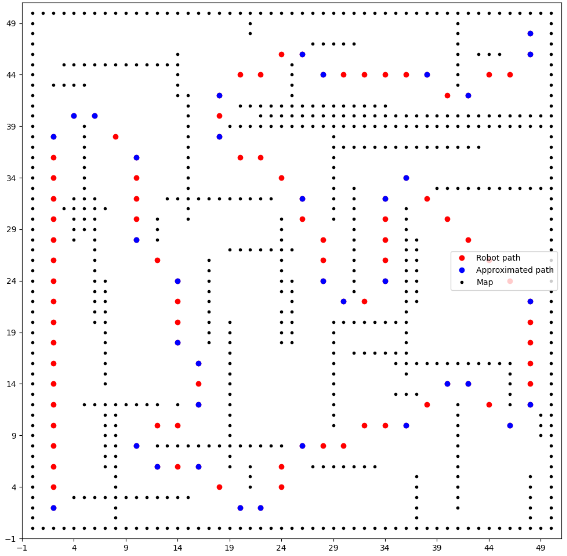}  
  \caption{Search produced scenario}
  \label{fig:sub-room2}
\end{subfigure}
\caption{Examples obtained robot navigation maps}
\label{fig:fig_maps}
\end{figure}

In Fig.\ref{fig:fig_maps} we show examples of the generated scenarios, i.e., rooms with obstacles obtained by random generation Fig. \ref{fig:sub-room1} and with AbmieGen Fig. \ref{fig:sub-room2}.
In Fig. \ref{fig:sub-room1} the length of the robot path towards the goal is 78.76 meters, while in the Fig. \ref{fig:sub-room1}  it is 202.36 meters. Evidently, the second scenario poses a more challenging navigation environment for the robot, than the first scenario.
The video demonstration of the fault revealed for the robot model in the Player/Stage environment can be found via the link: \url{https://figshare.com/s/7208f6d5ce19e1476474}.

 \subsection{Lane keeping assist system  case study{\label{sec:case2}}}
 
 Self-driving cars have a perspective of becoming a part of our lives in the near future. These systems are safety-critical and should be well tested to avoid unwanted consequences. Running the simulations in the virtual environments can reveal the possible faults of their control algorithms. %at the early design stage.
 
In this case study, we  generate virtual roads to test car Lane Keeping Assist System (LKAS). \Dima{The  ego-car, i.e., the test subject, should follow the lane of a given trajectory. The testing approach goal is to generate a valid road topology, that forces the ego-car to drive off its lane}. A number of tools were suggested for automatic generation of virtual roads, such as DeepJanus \cite{RiccioTonella_FSE_2020} and AsFault \cite{gambi2019automatically}. This year, four tools, such as Frenetic, Deeper, Swat, and GA-Bézier were presented at the SBST2021 tool competition \cite{report}. The SWAT tool is the submission of the random generator based implementation of our approach for virtual road generation.
 \subsubsection{System under test description.}
\begin{figure}[ht]
\includegraphics[scale=0.23]{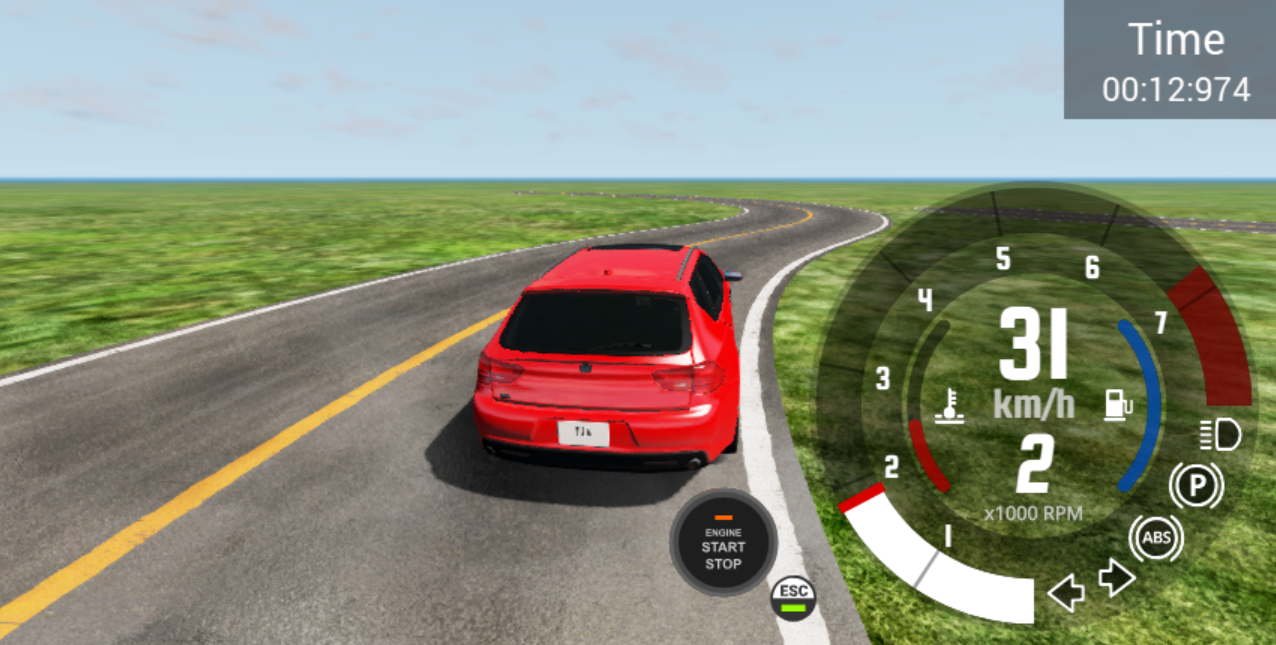}
\centering
\caption{The screenshot from a BeamNG simulation environment}
\centering
\label{fig:BeamNG}
%\vspace{-5mm}
\end{figure}

For simulating the car and the environment, we used the simulation pipeline initially provided by \cite{RiccioTonella_FSE_2020} and adapted for the SBST2021 tool competition.  %It was also used at SBST2021 competition.
This environment uses the BeamNG.tech driving
simulator \cite{beamng}, a freely available research-oriented version of the
commercial game BeamNG.drive (see Fig. \ref{fig:BeamNG}). The  test subject is the builtin
driving agent, BeamNG.AI. The car controller adopts a behavioural cloning approach, i.e., the deep learning component (DN)
learns a direct mapping from the sensor camera input
to the steering angle value to be passed to the actuators \cite{7410669}.

\subsubsection{Problem representation}
In this case study, the test scenario is a flat road
surrounded by plain green grass with the fixed weather conditions: sunny clear day. The road layout (i.e., number and width of lanes) is fixed and consists of two lanes.

\Dima{We divide the road into $m$ road segments. Each environment element $E_i$ corresponds to one road\Dima{segment.}}To describe the road\Dima{segment}we define three attributes:
the type of the road $A_1$: going straight, turning right and turning left. $A_2$: the length of the straight road segment, and $A_3$: the angle of the turn of the curved segment. The attributes representation is shown in Table \ref{tab:tab_veh}.
The test scenario contains 3 rows ($N=3$) and a variable number of columns $M$, depending on how many road segments fit in a map. In our scenarios and  at SBST competition, the map size was 200 x 200 m.

\begin{table}[ht]
\begin{center}
\caption{Attribute types for vehicle problem}
\label{tab:tab_veh}
\scalebox{0.85}{
\begin{tabular}{|c|c|c|} 
\hline 
$A_1,\; road \; segment \;type$ & $A_2, \; straight \; road \; length$ & $A_3, \; road \;  turn \; angle$\\
\hline
["straight", & [5, 6, ..., 50] & [5, 10,..., 85 ]\\
"turn left", & & \\
"turn right"] &  & \\
\hline
\end{tabular}}
\end{center}
\end{table}
\begin{table}[h]
\begin{center}
\caption{Example of individual representation for the second case study}
\label{tab:individ3}
\scalebox{0.85}{
\begin{tabular}{|c|c|c|c|c|c|} 
%\hline
%\textbf{Original value}& \textbf{Encoded value} &  \textbf{Original value} & \textbf{Encoded value}\\ 
\hline 
 & $E_1$ & $E_2$ & $E_3$ & $E_4$ & $E_5$\\
\hline
$A_1, \; road \; type$ & 0 & 1 & 1 & 2 & 0\\
\hline
$A_2, \; straight \; road \; length $ &  15 & - & - & - & 5 \\
\hline
$A_3, \; turning \; angle $ & 0 & 60 & 60 & 75 & 0 \\
\hline
\end{tabular}}
\end{center}
\end{table}

\begin{figure}[h]
\includegraphics[scale=0.4]{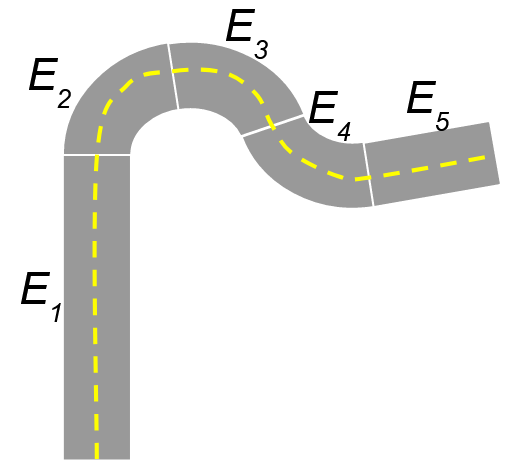}
\centering
\caption{An example of the test case represented by an individual in table \ref{tab:individ3}}
\centering
\label{fig:individ3}
%\vspace{-5mm}
\end{figure}

\Dima{An example of individual encoding that represents a road topology illustrated in in Fig.\ref{fig:individ3} is shown in Table \ref{tab:individ3}. Each environmental element, $E_i, \; i \in [1, 5]$, represent one road segment. The road segments are encoded as follows: 0 - for straight type, 1 - turning right and 2 - turning left}.

\begin{figure}[h]
\includegraphics[scale=0.41]{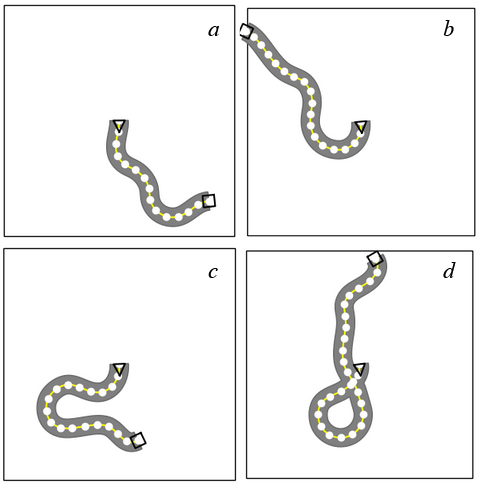}
\centering
\caption{Examples of valid (\textit{a}) and invalid roads: (\textit{b}) - out of bounds, (\textit{c}) - too sharp, (\textit{d}) - intersecting}
\centering
\label{fig:fig_roads_v}
%\vspace{-5mm}

\end{figure}

The test cases have the following \Dima{restrictions}:
the roads \Dima{cannot} be too sharp, \Dima{cannot} intersect and should not go out of the map bounds.
Examples of valid and invalid roads are shown in Fig. \ref{fig:fig_roads_v}.

\begin{comment}
\begin{figure}
\begin{subfigure}{.2\textwidth}
  \centering
  % include first image
  \includegraphics[scale=0.32]{images/valid2.PNG}  
  \caption{Valid road}
  \label{fig:sub-first}
\end{subfigure}
\begin{subfigure}{.2\textwidth}
  \centering
  % include second image
  \includegraphics[scale=0.32]{images/inv1.PNG} 
  \caption{Road is out of bounds}
  \label{fig:sub-second}
\end{subfigure}

\begin{subfigure}{.2\textwidth}
  \centering
  % include third image
  \includegraphics[scale=0.32]{images/inv2.PNG}  
  \caption{Road is too sharp}
\end{subfigure}
\begin{subfigure}{.2\textwidth}
  \centering
  % include fourth image
  \includegraphics[scale=0.32]{images/inv3.PNG}  
  \caption{Road is self-intersecting}
\end{subfigure}
\caption{Examples of valid and invalid roads}
\label{fig:fig_roads}
\end{figure}

\end{comment}

\subsubsection{Fitness function definition}
\begin{figure}[ht]
\includegraphics[scale=0.28]{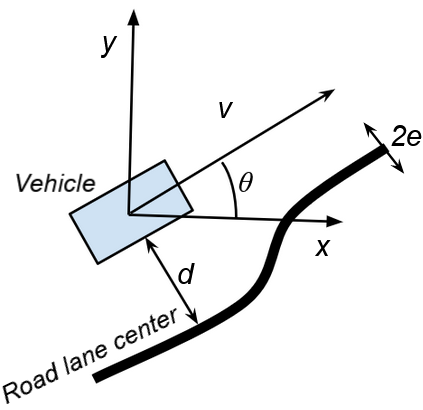}
\centering
\caption{The simplified car model parameters}
\centering
\label{fig:model_}
%\vspace{-5mm}
\end{figure}

To calculate the test scenario fitness we need to create the simplified model of the car. Similarly to the thermostat problem, we built the car model from the first principles as the car movement can be described by a well known car kinematic model \cite{laumond1998guidelines}.
To describe the car movement we use the equations from \cite{alur2015principles}, see Fig.\ref{fig:model_}. To keep the car close to the lane center we adopt  Stanley control \cite{hoffmann2007autonomous}.

In the equations below, $x, y$ - are the current coordinates of the car on the map, $\theta$ is the angle between car direction and a reference plane, $a$ and $b$ - constants, corresponding to velocity value, $d$ - the distance of the car from the closest point on the road. When $d$ is smaller than a certain threshold $e$, the car goes straight, when $d$ is larger than $e$ - the car turns either left or right. The turn angle is adopted depending on the car speed and the deviation from the road lane center.
%Also, when the car goes straight we increase the car speed by a small value $\alpha$, when the car is turning, the speed is decreased by a small value $\beta$.
%The   longitudinal   velocity   is   assumed   as constant.

Therefore we have the following fine-tunable parameters:
$k$, $\alpha$, $\beta$ and the initial speed $\nu_0$.
In order to fine tune the parameters, we created a dataset with the road points and the corresponding car model path $S$ recorded by the simulator while executing the scenarios. Then we compared the outputs of our model with the simulated car path using such metric as a "Hausdorff distance". A similar metric, Frechet distance, was used in \cite{kluck2021automatic} to compare the similarity between roads. The goal was to minimize the Hausdorff distance.
To perform the optimization we use the sci-py Nelder-Mead algorithm implementation. However, other optimization algorithms can be used, such as genetic algorithms.
The set of the parameters that indicated the lowest average Hausdorff distance of 13.74 is shown in Table \ref{tab:veh_param}.
\begin{table}[ht]
\begin{center}
\caption{Fine tuned values for the surrogate model}
\label{tab:veh_param}
\scalebox{0.95}{
\begin{tabular}{|c|c|c|c|} 
\hline 
$\nu_0$ & $k$ & $\alpha$ & $\beta$\\
\hline
7 & 3.5 & 0.3 & 0.1\\
\hline
\end{tabular}}
\end{center}
\end{table}
\begin{gather}
    \dot{x} = \nu \cdot cos\theta \\
    \dot{y} = \nu \cdot sin\theta \\
    %\dot{\theta} = \tan^{-1} (\frac{k \cdot d(t)}{\nu(t)})
      \dot{\theta} =
    \begin{cases}
      \tan^{-1} (\frac{k}{\nu(t)}) & \text{$if \; d < -e$}\\
      -\tan^{-1} (\frac{k }{\nu(t)}) & \text{$if \; d > e$}\\
      0 & \text{$if -e \leq d \leq e$}
    \end{cases}\\
    \dot{\nu} =
       \begin{cases}
      -\alpha & \text{$if \; d < -e, d > e$}\\
      \beta & \text{$if -e \leq d \leq e$}\\
    \end{cases}
\end{gather}

\begin{figure}[ht]
\includegraphics[scale=0.27]{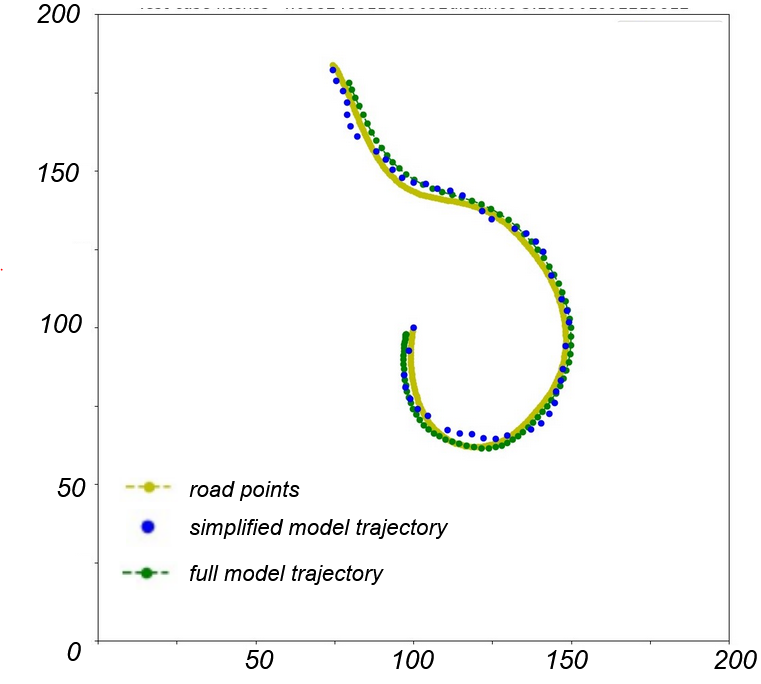}
\centering
\caption{The simplified and full car model trajectory given the same road points}
\centering
\label{fig:model_hau}
%\vspace{-5mm}
\end{figure}
In Fig.\ref{fig:model_hau} \Dima{we depict} the surrogate (blue points) and the full model (green points) follow the interpolated road points (yellow). The  Hausdorff distance between the two roads is 5.153.

\Dima{In this case study the requirement for the vehicle is to stay within the road lane bounds and not go out of the lane for more than a defined threshold, proportional to the area of the vehicle outside the lane. For instance, the threshold of 0.5 means, that the car is considered to be out of the lane if more than the half of the vehicle lies outside the lane. In our case study we use a threshold of 0.85.}

Finally, as the fitness function, $F_{1veh}$, we used the maximum deviation $d$ from the lane center after executing the test case, as in \cite{gambi2019automatically} and \cite{riccio2020model}.\Dima{The test cases with the highest value of the fitness function are most likely to violate the requirement, as they force the car to deviate further from the lane centre.} For the test cases that don’t meet the restrictions, $F_{1veh}$ is set to 0.  The second fitness function was calculated according to (\ref{jaccard}).
\subsubsection{Genetic algorithm configuration.}
We used the following GA and NSGA-II configurations:
population size: 500, 
number of generations: 200,
mutation rate: 0.4,
crossover rate: 1,
algorithm type: generational,
number of evaluations: 100 000.

We are using a higher population size, rather than the bigger number of generations, as from our experience, with bigger population the results were more consistent across different runs. 
The average time to run 100000 evaluations was 1522.405 sec for GA and 1380.66 sec for NSGA-II.
\subsubsection{Scenario creation}
To create the road points $p_1-p_7$, used to generate the road, we applied affine transformations to the initial vector $\nu_1$, according to the road types specified in the generated scenario, i.e., "straight N meters", "turn right/left N degrees". Our approach is described in a more detail in \cite{swat_tool}. 
\begin{figure}[ht]
\includegraphics[scale=0.3]{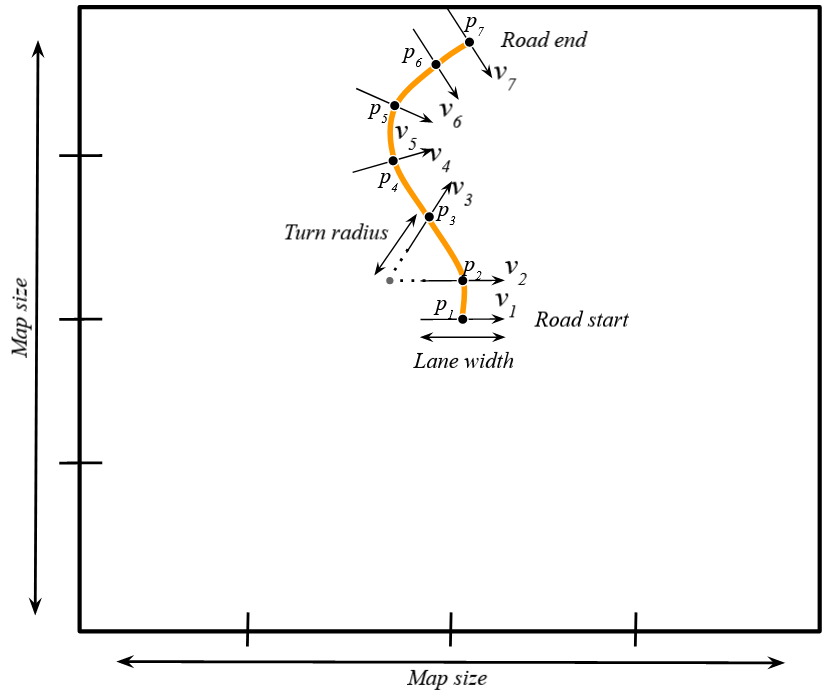}
\centering
\caption{The road points (black) generated after applying affine transformations to the initial vector $\nu_1$}
\centering
\label{fig:swat_gen}
%\vspace{-5mm}
\end{figure}

\begin{figure*}[h!]
\begin{subfigure}{.45\textwidth}
  \centering
  % include third image
  \includegraphics[scale=0.3]{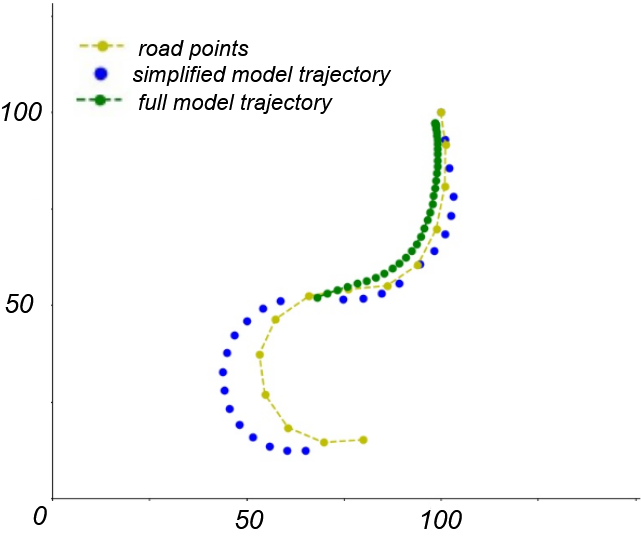}  
  \caption{Scenario forcing the car to drive off the lane}
\end{subfigure}
\begin{subfigure}{.45\textwidth}
  \centering
  % include fourth image
  \includegraphics[scale=0.27]{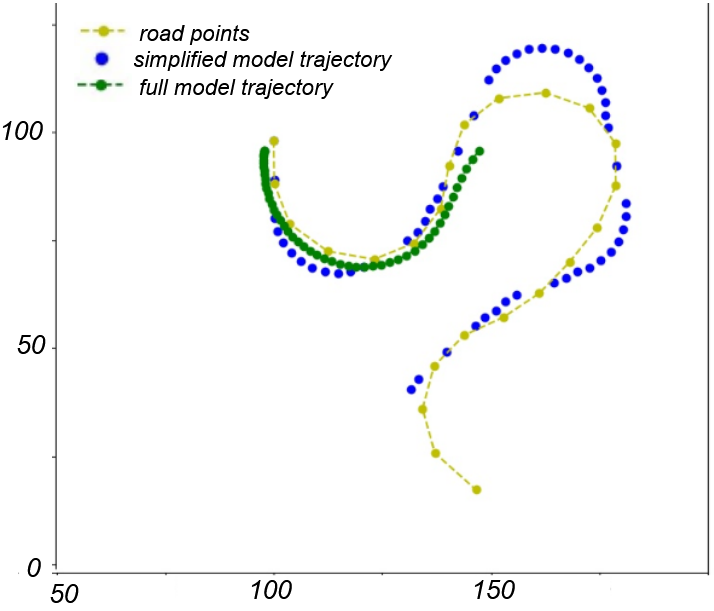}  
  \caption{Scenario forcing the car to drive off the lane}
\end{subfigure}
\caption{Examples of fault revealing scenarios for vehicle lane keeping assist system}
\label{fig:fig_roads}
\end{figure*}

For example, to obtain $\nu_2$ we moved the $\nu_1$ paralelly N meters ("straight N meters"). To obtain $\nu_3$ we turned $\nu_2$ N degrees anticlockwise ("turn left N degrees").
In the Fig. \ref{fig:fig_roads} we show examples of the generated test cases, that forced the car to go out of the lane. The yellow points correspond to the road lane center, the blue points - to the surrogate model path, green points - the full model path. When the virtual car went out of the lane bounds, the simulation recording stopped, therefore we see the full model path only for the part of the road.
The video demonstration of the failure for the car model can be found via the link: \url{https://figshare.com/s/b4a096f0a66e0abbe7b1}.

 \section{Experimental evaluation{\label{sec:eval}}}

\subsection{Research questions}
We evaluate our approach using the three test case generation case studies described above. For each of the case studies we answer the following research questions.

\textbf{RQ1. (Comparing random, single-objective, and multi-objective search)}. \textit{%Does the multi-objective search, accounting for fault revealing power and diversity, perform better in maximizing the fault revealing power of the scenarios than the single-objective and random search
Considering the single and multi objective versions of AmbieGen as well as the random search, which configuration produces the test scenarios with the higher fault revealing power given the same time budget}?

\textbf{Motivation:} Firstly we would like to know if the use of evolutionary search is beneficial and allows to produce better solutions, than simple random search. Next, we  want to know if adding additional fitness function for diversity allows to find solutions with a\Dima{higher fault revealing power. We expect AmbieGen MO to produce solutions with the fault revealing power, i.e., $F_1$ fitness function value at least as high as for the solutions of AmbieGen SO.}
Previous works on novelty search \cite{de2001reducing}, \cite{mouret2011novelty} have shown that adding a fitness function for diversity may increase the convergence speed. 

\textbf{Experiment design:}
We give the\Dima{same number of evaluations}to all the three algorithms and compare the average $F_1$ fitness function value of the best solutions found. We repeat the measurements 30 times.

\textbf{RQ2. (Comparing diversity of the solutions found by the single-objective and multi-objective search).} \textit{To what extent the diversity of the solutions found by the multi objective AmbieGen configuration is higher than the diversity of the single objective configuration solutions?} 

\textbf{Motivation:} This research question is aimed to quantify the difference between the diversity of the solutions produced by AmbieGen So and AmbieGen MO. We expect the AmbieGen MO to produce more diverse scenarios.

 %In the test generation problem, after running the algorithm, it is often preferable to obtain a test suite, rather than a single test case.
\textbf{Experiment design:}
Given the same time budget, we compare the average diversity of the best 10 solutions found by the single-objective algorithm and the average diversity of the Pareto optimal solutions found by the multi-objective algorithm. We repeat each measurement 30 times.

For the autonomous robot and lane keeping assistant case study we also answer the following question:

\textbf{RQ3. (Comparing our AmbieGen with the available baselines)} \textit{To what extent does our approach perform better in generating test scenarios for the full model in comparison with the available baselines?} 

\textbf{Motivation:}  This research question is aimed to quantify the effectiveness of AmbieGen in the number of revealed failures for the full models used in simulations.  %We answer this question only for the last two case studies as for the first cases study the results were obtained from the real system data. Therefore we 

\textbf{Experiment design:}

\textit{Autonomous robot case study.}
To the best of our knowledge, there are no available test generation baselines for the  autonomous robot system. Therefore we compare the generated scenarios with the random search by giving the same time budget of two hours and executing the generated environments in the robotic simulator. We repeat the experiment 30 times.

\textit{Lane keeping assist system.}
For the lane keeping assist system, we compare AmbieGen with the open-source approach that showed the best results in the SBST2021 tool competition \cite{report},  i.e., Frenetic tool \cite{frenetic}. In the competition the same test evaluation pipeline was provided to all the participants. It allowed to compare the generated test cases for the number of faults revealed (forcing the ego-car to go out of the lane), the diversity of the revealed faults and the proportion of valid test cases. We perform the same 2 hour experiment as in the competition, averaging the results over 30 runs. 

For all the research questions, to confirm the statistical significance of the results we performed a two-tailed non-parametric Mann-Whitney U test and measured the effect size using the non-parametric measure Cliff's delta. 
We ran all the experiments on a PC running
Microsoft Windows 10 Home and featuring a quad-core AMD Ryzen 7 4800HS CPU @ 2.90 GHz, 16 GB of Memory, and
an NVidia GeForce GTX 1660 GPU @ 6GB.

%\subsection{Experimental protocol}

\subsection{Results}
%\Foutse{Please present the results organized by problems, not by RQs...you are repeating the rq structure too much! Also discuss the results of each problem here and not in the discussion section.}

\begin{figure}[h!]
\includegraphics[scale=0.47]{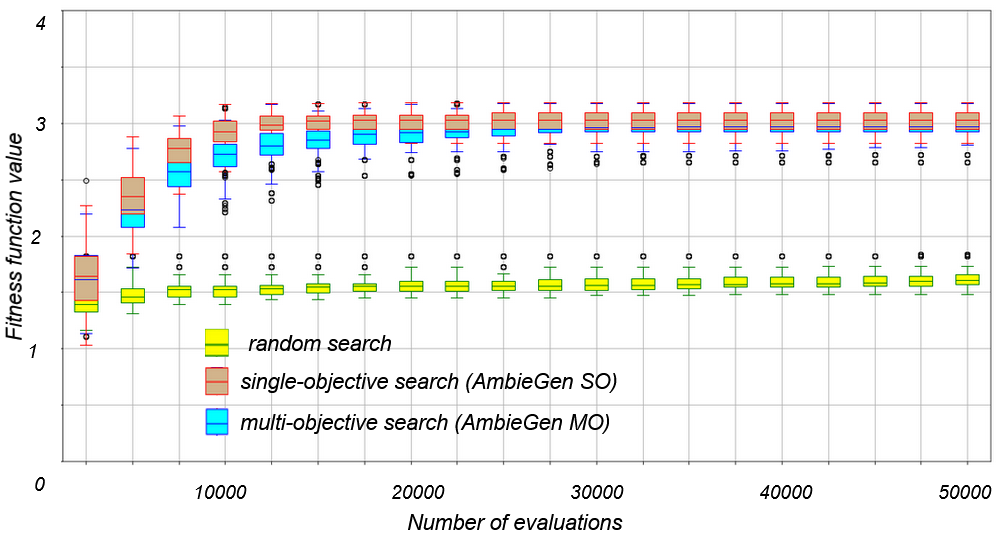}
\centering
\caption{Best fitness function value over evaluations for the thermostat case study}
\centering
\label{fig:therm_res}
%\vspace{-5mm}
\end{figure}

\begin{figure}[h!]
\includegraphics[scale=0.48]{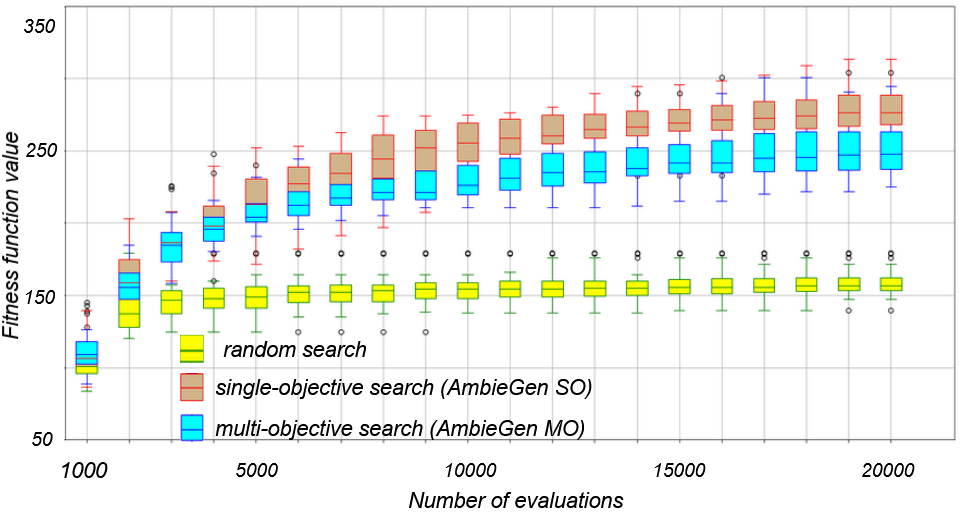}
\centering
\caption{Best fitness function value over evaluations for the autonomous robot case study}
\centering
\label{fig:rob_res}
%\vspace{-5mm}
\end{figure}

\begin{figure}[h!]
\includegraphics[scale=0.56]{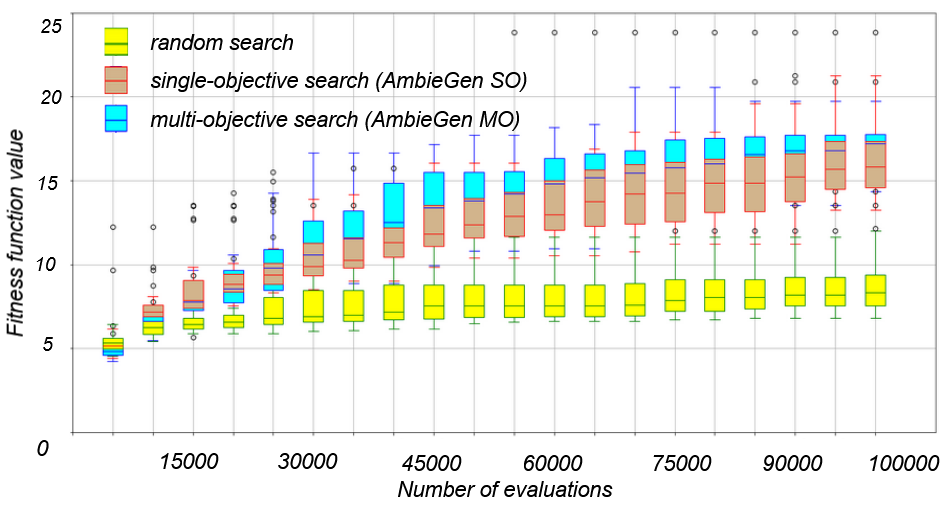}
\centering
\caption{Best fitness function value over evaluations for the lane keeping assistant case study}
\centering
\label{fig:veh_res}
%\vspace{-5mm}
\end{figure}

\textbf{RQ1.(Comparing random, single-objective, and multi-objective search)} In the Fig. \ref{fig:therm_res},  Fig. \ref{fig:rob_res}, and Fig. \ref{fig:veh_res}, we present the best fitness value found over generations by Random search (green boxplots), AmbieGen SO  (red boxplots) and AmbieGen MO (blue boxplots) averaged over 30 runs for the three problems. We considered the fitness function accounting for the fault revealing power and described in Equation (\ref{f1}).

We  compare the fitness function values found after the allowed number of evaluations with a two-tailed non-parametric Mann-Whitney U test. The obtained p-values and effect sizes of the problems are shown in the Tables \ref{tab:therm_p}, \ref{tab:rob_p}, and \ref{tab:veh_p}, respectively.

\textit{Thermostat case study.}
From Fig. \ref{fig:therm_res} we can see that on average random search (yellow) converges to values of -1.608, while AmbieGen SO (red) and AmbieGen MO (blue) find the solutions with twice higher fitness value of -3.  Statistical tests confirm that AmbieGen outperforms the random search with p < 0.01.
We can observe that on average the SO converges faster than MO, however, the difference between the converged values is negligible. %On average, AmbieGen SO requires lower number of evaluations to converge.

\begin{table}[!ht]
\begin{center}
\caption{Results of two-tailed non-parametric Mann-Whitney U test and Cliff's delta effect sizes for the thermostat case study}
\label{tab:therm_p}
\scalebox{0.85}{
\begin{tabular}{|c|c|c|c|} 
\hline 
       & $SO \: (GA)$ & $MO \: (NSGA2)$ & $Random$ \\
\hline
$SO \: (GA)$  &       &         &           \\
\hline
$MO \: (NSGA2)$ &  $p = 0.378$  &          &           \\
        &  $0.133, negligible$  &         &           \\
\hline
$Random$ &  $p < 0.01$  &      $p < 0.01$    &    \\
         &  $1, large$  &      $1, large$    &     \\
\hline
\end{tabular}}
\end{center}
\end{table}

\textit{Autonomous robot case study.}
After 20000 evaluations, on average, random search  produced solutions with the highest fitness value of -158.89. AmbieGen outperforms the random search, with the SO configuration producing 42 \% fitter solutions of -278.2 and the MO configuration producing solutions of -251.6 fitness value. Given the same time budget, AmbieGen SO produces almost 10 \% fitter solutions than AmbieGen MO.

\begin{table}[h!]
\begin{center}
\caption{Results of two-tailed non-parametric Mann-Whitney U test and the Cliff's delta values for the autonomous robot case study}
\label{tab:rob_p}
\scalebox{0.85}{
\begin{tabular}{|c|c|c|c|} 
\hline 
       & $SO \: (GA)$ & $MO \: (NSGA2)$ & $Random$ \\
\hline
$SO \: (GA)$ &      &         &           \\
\hline
$MO \: (NSGA2)$ & $p < 0.01$  &      &           \\
        & $0.772, large$  &          &           \\
\hline
$Random$ &  $p < 0.01$  &      $p < 0.01$    &    \\
&  $1, large$  &      $0.978, large$    &     \\
\hline
\end{tabular}}
\end{center}
\end{table}

\textit{Lane keeping assistant case study.}
In 100000 evaluations random search produced scenarios with the average highest fitness of -9. AmbieGen MO and SO produced almost 50 \% fitter solutions of -17 and -16, respectively. There was no statistical difference between the best solutions of SO and MO.

Overall, We can see that for all the problems, AmbieGen finds on average from 40 \% to 50 \% better solutions, than the random search. AmbieGen SO and AmbieGen MO show no statistical difference in the produced solutions for the thermostat and lane keeping assistance problem. For the autonomous robot problem, the AmbieGen SO produces better solutions with a large effect size given 20000 evaluations.

\begin{table}[h!]
\begin{center}
\caption{Results of two-tailed non-parametric Mann-Whitney U test and the Cliff's delta values for lane keeping assistant case study}
\label{tab:veh_p}
\scalebox{0.85}{
\begin{tabular}{|c|c|c|c|} 
\hline 
       & $SO \: (GA)$ & $MO \: (NSGA2)$ & $Random$ \\
\hline
$SO \: (GA)$  &       &         &           \\
\hline
$MO \: (NSGA2)$ &  $p = 0.0501$  &        &           \\
        &  $0.295, small$  &      &           \\
\hline
$Random$ &  $p < 0.01$  &      $p < 0.01$    &    \\
         &  $0.877, large$  &   $0.886, large$    &    \\
\hline
\end{tabular}}
\end{center}
\end{table}

\begin{tcolorbox}[colback=white]
%SO and MO outperform the random search in all case studies with a "large" effect size. SO and MO show no statistical difference in the fitness of the produced scenarios for the thermostat and lane keeping assist problems. Given the same time budget the SO produced fitter individuals with a "large" effect size for the autonomous robot problem. Therefore we conclude SO is more likely to converge to a better fitness function given the same time budget.
%AmbieGen produces better solutions than the random search with a "large" effect size for all case studies. Single objective configuration tends to find the fitter solutions faster, than the multi-objective configuration.
AmbieGen SO produced scenarios with highest fault revealing power for the autonomous robot case study. For the thermostat and LKAS case studies the difference in the solution fitness of  AmbieGen SO and MO was negligible. Overall, AmbieGen outperforms the random search with "large" effect size in all case studies.
\end{tcolorbox}

\begin{figure}[h!]
\includegraphics[scale=0.5]{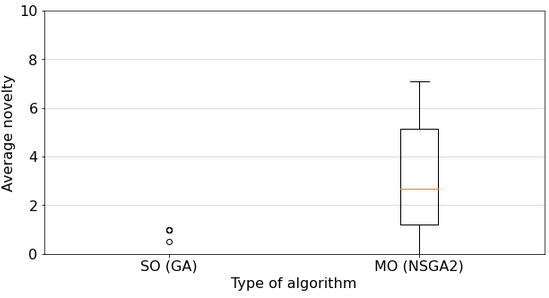}
\centering
\caption{Diversity of the test cases in the last generation for the thermostat case study}
\centering
\label{fig:div_therm}
%\vspace{-5mm}
\end{figure}

\begin{figure}[h!]
\includegraphics[scale=0.5]{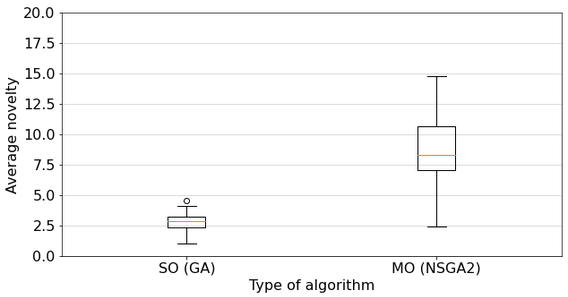}
\centering
\caption{Diversity of the test cases in the last generation for the autonomous robot case study}
\centering
\label{fig:div_rob}
%\vspace{-5mm}
\end{figure}

\begin{figure}[h!]
\includegraphics[scale=0.5]{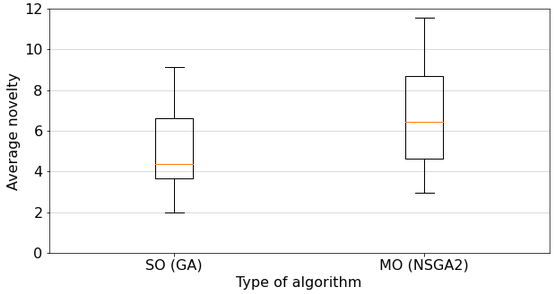}
\centering
\caption{Diversity of the test cases in the last generation for the lane keeping assistant case study}
\centering
\label{fig:div_car}
%\vspace{-5mm}
\end{figure}

\textbf{RQ2.(Comparing diversity of the solutions found by the single-objective and multi-objective search).} In Fig. \ref{fig:div_therm}, Fig.\ref{fig:div_rob}, and Fig. \ref{fig:div_car}, we compare how diverse are the produced solutions by AmbieGen SO and AmbieGen MO.

%We select 10 fittest individuals from the population and calculate the average of the diversity value between each pair of individuals. We calculate the diversity according to (\ref{jaccard}).

For SO, we select 10 fittest individuals and compute the diversity according to (\ref{jaccard}) between each pair of individuals. We report the average value.
For NSGA-II, we compute the diversity (\ref{jaccard}) between each pair of Pareto optimal solutions. The size of the Pareto front was 7 individuals on average. All the solutions in the Pareto front have a fault revealing ($F_1$) fitness function value higher than a certain fault-revealing threshold, established by the developer.

For all the problems, the two-tailed non-parametric Mann-Whitney U test confirmed that AmbieGen MO produces more diverse solutions, than AmbieGen SO.
For the smart thermostat problem, the MO scenarios are more diverse with a p-value smaller than 0.01 and a "large" effect size of 0.852.
For the autonomous robot problem, AmbieGen MO scenarios are more diverse with a p-value smaller and a "large" effect size of 1.
For the lane keeping assist system, MO scenarios are more diverse with a p-value of 0.0109 ($p \leq 0.05$) and  a "medium" effect size of 0.383.

\begin{tcolorbox}[colback=white]
In all the considered problems AmbieGen MO produced more diverse test cases: with "large" effect size for thermostat and robot case\Dima{studies}and "medium" effect size for the LKAS case study.
\end{tcolorbox}

\begin{tcolorbox}[]
\textit{(RQ1, RQ2 summary:)} AmbieGen MO can find scenarios of the same quality as AmbieGen SO and better scenarios with a large effect size than the random search. Moreover, AmbieGen MO produces a more diverse set of scenarios, than AmbieGen SO. %Therefore for the future evaluations we are using the 
Overall, we recommend using the AmbieGen MO configuration.
%In the three considered case studies AmbieGen MO finds the individuals of similar quality as AmbieGen SO, producing a more diverse set of test scenarios. Therefore, we recommend using the multi objective configuration of AmbieGen. For the evaluations of AmbienGen in RQ3 we are using the AmbieGen MO configuration. 
\end{tcolorbox}
\textbf{RQ3.(Comparing AmbieGen with the available baselines).}

\begin{figure}[h!]
\includegraphics[scale=0.48]{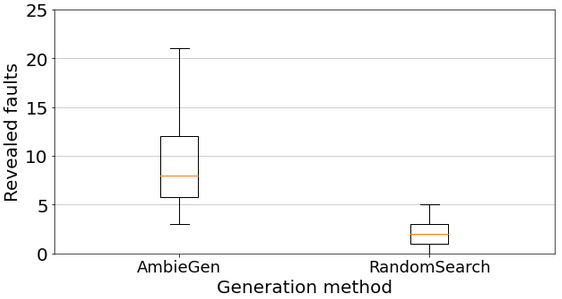}
\centering
\caption{Number of faults revealed for the autonomous robot}
\centering
\label{fig:obe_robo}
%\vspace{-5mm}
\end{figure}

\textbf{Autonomous robot case study}. In this subsection we compare the number of faults revealed by the NSGA-II configuration of AmbieGen (AmbieGen MO)  and the random search.
We created a scenario evaluation pipeline, where firstly a two hour budget is given to produce the scenarios. Then all the scenarios are passed to the simulator and executed. The daemon script monitors the execution and reports a failure when the robot stalls and does not reach a goal. We repeated the experiment 30 times in both configurations. You can see the average number of failures detected  in Fig. \ref{fig:obe_robo}.
AmbieGen produced on average 9 failures in two hours, in comparison to the 2 failures of random search. AmbieGen outperforms the random search with a p-value less than 0.01 and a large effect size of 1.

\textbf{Lane keeping assist system case study}. In this subsection we report results of evaluating  AmbieGen MO (AmbieGen) and the Frenetic tool (Frenetic). In addition we evaluate the random search (RS) configuration of AmbieGen and the AmbieGen MO configuration based on the full model (Full).

%Previously we submitted the random generator based implementation of our road generation approach to SBST2021 tool competition under the name of "SWAT" tool. It produces the virtual roads  randomly, with no search. It generated a high proportion of the valid scenarios, however revealed a small number of failures i.e on average . 

For AmbieGen we used a simplified configuration for virtual road generation, where only, 5,100 evaluations are performed (population size 100, number of generations 200, number of offsprings - 25) in order to produce more test scenarios given a limited time budget.
%population  size:  90,  number  of  generations: 200, mutation rate: 0.4, crossover rate: 1, number of offsprings: 25, number of evaluations: 5090.
%We used the simplified version to produce more test cases given a limited time budget for evaluation (2 hours). 
We gave the same time budget (5,090 evaluations) for the random search to produce the solutions.

Finally, for the full model we used a configuration previously suggested by Gambi et al. \cite{gambi2019automatically} for Asfault tool, that also uses the full model to guide the search. In this configuration the population size is 25, number of offsprings is 4 and the number of generations is limited by the time budget, i.e, two hours.

Approaches were evaluated using the SBST2021 code pipeline \cite{riccio2020model}, that integrates the test generators with the BeamNG simulator by validating, executing, and evaluating the generated test cases.  We executed the SBST21 2 hour experiment, where the\Dima{failure}is revealed when 0.85 percent of the car area goes out of the lane. Also, the driving agent travels up to 70 Km/h.

\begin{figure}[h!]
\includegraphics[scale=0.75]{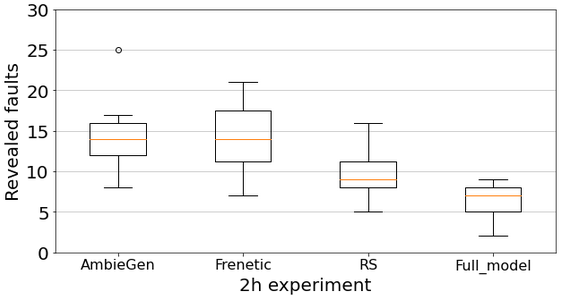}
\centering
\caption{The number of revealed faults}
\centering
\label{fig:obe_car}
%\vspace{-5mm}
\end{figure}

\begin{figure}[h!]
\includegraphics[scale=0.75]{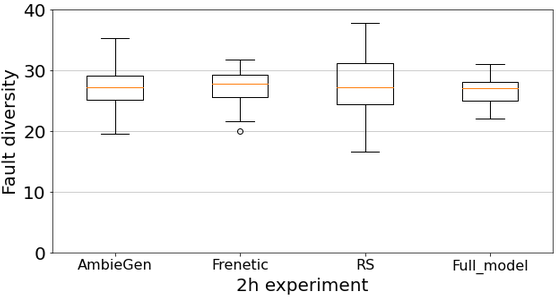}
\centering
\caption{The diversity of the revealed faults}
\centering
\label{fig:spars_car}
%\vspace{-5mm}
\end{figure}

\begin{figure}[h!]
\includegraphics[scale=0.55]{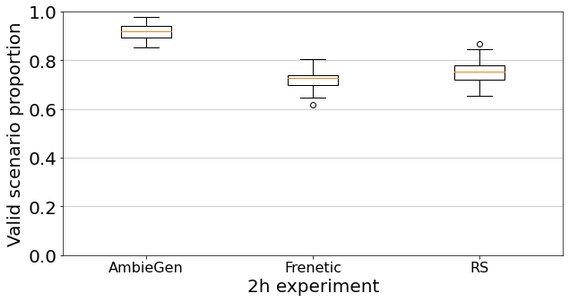}
\centering
\caption{The proportion of the valid test cases}
\centering
\label{fig:valid_car}
%\vspace{-5mm}
\end{figure}

The test cases are compared in terms of the number of faults Fig. \ref{fig:obe_car}, the diversity of the faults Fig. \ref{fig:spars_car}, and the proportion of the valid test cases Fig. \ref{fig:valid_car}.
The corresponding statistical test and effect size measures (Cliff's delta) are shown in the Tables \ref{tab:obe}, \ref{tab:spars} and \ref{tab:ratio}.

\begin{table}[h!]
\begin{center}
\caption{ Mann-Whitney test p value and Cliff's delta for the number of faults}
\label{tab:obe}
\scalebox{0.85}{
\begin{tabular}{|c|c|c|c|c|} 
\hline 
          & $AmbieGen$  & $Frenetic$ & $Full$ & $RS$ \\
\hline
$AmbieGen$     & &            &           &        \\
\hline
$Frenetic$ &$p = 0.917$            &&             &         \\
           &$0.0166, \; negligible$&            &           &         \\
\hline
$Full$   &$p < 0.01$            &$p < 0.01$&   &       \\
           &$0.996, \; large$     &$0.991, \; large$&    &       \\
\hline
$RS$     &$p < 0.01$       &$p < 0.01$&    $p < 0.01$          &   \\
         &$0.653, \; large$&$0.578, \; large$&  $0.951, \; large$            &  \\
\hline
\end{tabular}}
\end{center}
\end{table}
In terms of the number of the revealed faults both, AmbieGen and Frenetic, statistically outperform the random search and the full model based search. Out of 30 runs, on average, AmbieGen and Frenetic produce almost equal amount of faults, i.e., 14.

\begin{table}[h!]
\begin{center}
\caption{ Mann-Whitney test p value and Cliff's delta for the fault sparsity}
\label{tab:spars}
\scalebox{0.85}{
\begin{tabular}{|c|c|c|c|c|} 
\hline 
          & $AmbieGen$  & $Frenetic$ & $Full$ & $RS$ \\
\hline
$AmbieGen$     & &            &           &        \\
\hline
$Frenetic$ &$p = 0.897$            &&             &         \\
           &$0.020, \; negligible$&            &           &         \\
\hline
$Full$   &$p = 0.0889$            &$p = 0.0998$&    &       \\
           &$0.3238, \; small$     &$0.315, \; small$&    &       \\
\hline
$RS$     &$p = 0.912$       &$p = 0.794$&    $p = 0.147$          &   \\
         &$0.018, \; negligible$&$0.042, \; negligible$&  $0.285, \; small$            &  \\
\hline
\end{tabular}}
\end{center}
\end{table}

Concerning the diversity of the revealed faults, all the approaches have similar performance and do not show a statistically significant difference.

\begin{table}[h!]
\begin{center}
\caption{ Mann-Whitney test p value and Cliff's delta for the proportion of valid cases}
\label{tab:ratio}
\scalebox{0.85}{
\begin{tabular}{|c|c|c|c|c|} 
\hline 
          & $AmbieGen$  & $Frenetic$ &  $RS$ \\
\hline
$AmbieGen$     & &            &                 \\
\hline
$Frenetic$ &$p < 0.01$            &&                      \\
           &$1, \; large$&            &                   \\
%\hline
%$Full$   &$p = 0.629$            &$p < 0.01$&    &       \\
%           &$0.084, \; negligible$     &$1, \; large$&    &       \\
\hline
$RS$     &$p < 0.01$       &$p = 0.011$&              \\
         &$0.997, \; large$&$0.386, \; medium$&              \\
\hline
\end{tabular}}
\end{center}
\end{table}
\begin{table}[h!]
\begin{center}
\caption{ Number of total, valid and invalid test cases}
\label{tab:tc_number}
\scalebox{0.85}{
\begin{tabular}{|c|c|c|c|} 
\hline 
          & $TCs$    & $Valid$ & $Invalid$ \\
\hline
$AmbieGen$     &$150$ &    $137.8$        &     $12.23$          \\
\hline
$Frenetic$ &$190.3$&$136.46$  &       $53.83$           \\
\hline
%$random$   &$223.57$&$206.21$&   $17.36$       \\
%\hline
$RS$    &$86.53$ &  $65.32$&    $21.21$            \\
\hline
\end{tabular}}
\end{center}
\end{table}

Another important factor was the proportion of valid test cases out of all the cases produced. From Table \ref{tab:ratio} we see that AmbieGen produces a statistically \Dima{larger} proportion of the valid test cases, than Frenetic and random search. For the full model, the invalid scenarios were assigned the fitness value of 0 and not submitted for evaluation.
In Table \ref{tab:tc_number} we also indicate the average number of the total produced test cases as well as the number of invalid and valid test cases. For the full number, initially the 25 individuals were produced that were later evolved by the search operators. SBST2021 code pipeline evaluates the test cases procedurally, i.e., as soon as the valid test case is produced it is executed. The new test case can only be produced, when the execution of the previous one stops. The scenario execution time depends on the generated road length, i.e., the longer the road, the more time the car will spend in the simulation. Therefore, we do not evaluate the approaches by the total number of the produced scenarios, as it depends not only on the efficiency of the algorithm, but also on the duration of the generated scenarios. The random search generates the lowest number of solutions as it only provides one solution after 5,090 evaluations. AmbieGen, on the contrary, provides around 7 solutions on average, corresponding to the search Pareto front after 5,090 evaluations.

\begin{tcolorbox}[colback=white]
AmbieGen reveals 9 failures in two hours, in comparison, random search could reveal only 2 failures for the robot case study.
Both AmbieGen and state of the art Frenetic tool revealed 14 failures in two hours. The revealed faults have similarly high diversity for both tools. AmbieGen outperforms Frenetic in the proportion of the valid generated scenarios. AmbieGen also outperforms the random search and the full model configuration in the number of revealed faults.
\end{tcolorbox}

% \section{Results{\label{sec:results}}}
%\input{results}
 \section{Discussion{\label{sec:disc}}}
%\Foutse{the questions should be answered in the result section and not in a discussion section! In discussion it is expected that you will provide a general discussion of your results and explain how impactful they are, how they affect/impact the practice and what future research avenues they are opening...etc}.%In this section we provide the answers to the formulated research questions, based on the results described in the previous section.

%Answering the first research question, SO and MO outperform the random search in all case studies with a "large" effect size. SO and MO show no statistical difference in the fitness of the produced scenarios for the thermostat and lane keeping assist problems. Given the same time budget the SO produced fitter individuals with a "large" effect size for the autonomous robot problem. Therefore we conclude SO is more likely to converge to a better fitness function given the same time budget.

%In our study we are using a novel technique to represent an environmet for a cyber-physical system. 
In this section we discuss the implications of our findings for research and the practice.\\
\textbf{Evolutionary algorithms for scenario generation.}
Evolutionary algorithms were proven to be effective, comparing to random generation, to create virtual environments for testing automotive systems in previous works such as \cite{ben2016testing}, \cite{gambi2019automatically}, \cite{RiccioTonella_FSE_2020}. The implementation of such algorithms to generate environments is rather challenging as the customized solution representation and search operators need to be developed.

In our work we extend the application of evolutionary search for environment generation to such domains as smart-homes and autonomous robots .
This work is the first stage in designing a framework for generating virtual environments, AmbieGen. We consider the complete virtual environment to be composed of separate elements. Each element is described with a fixed number of attributes. During the search we recombine the elements as well as their attributes. One of the advantages of such representation is the simplicity of implementation of initial population generation, crossover and mutation search operators. Therefore the developer only needs to consider a high level description of the problem and not concentrate on the design of search operators and solution representation. % In Asfault paper for example, one of the challenges was to merge the road segments after applying a crossover operator. 
By adding more attributes, the scenario complexity can be increased. For the smart-thermostat, for example, we can add such attributes as the humidity inside the room and temperature outside the room for each time period. For the autonomous robot - the terrain type and indicate the presence of other robots. For the car, for each road section  we can indicate the terrain type, the incline, the location of the other vehicles, etc. 

Overall, our study confirms the effectiveness of search based approaches for environment generation. %However, the design of the search algorithms is rather challenging. 
Our framework  is aimed to reduce the effort of the developers of evolutionary algorithms to test the autonomous CPS. We provide the structure of the solution representation  and search operators, which can be applied to generation of different types of environments. We provide examples of generating smart-thermostat schedules, maps with obstacles and virtual roads with search algorithm implementation based on Pymoo framework.

\textbf{Using simplified system models.}
We explore the possibility to use the approximated system models, rather than the full models to compute the fitness function. Evidently, full models can detect failures with a higher precision, however they  are more expensive to execute in terms of resources and time budget. For instance, the recommended requirements for running BeamNg simulator are 16 GB RAM, Nvidia\\ GeForce GTX 970 videocard and Intel Core i7-6700 3.4Ghz processor or better. 
Our evaluations have shown that the full model failures can be detected by an approximated model. Moreover, given the same time budget the search guided by the approximated model may reveal more faults, than when guided by the full model. We advocate for the development of more precise simplified CPS models and making them open source, so that they can be easily used by researchers to calculate the search fitness functions. 
The possibility to use the surrogate models was first suggested by \cite{menghi2020approximation}, however it was used only to generate the CPS inputs. In \cite{ben2016testing}  the approximated models were used to generate the environments, however no comparison with the full model configuration was provided.

\textbf{Challenges.} The challenging part of AmbieGen implementation is in evaluating the test scenario fitness. It consists of two stages: the first is to convert the high level description matrix $TC$ to the environment configuration. For the smart thermostat we needed to convert the $TC$ matrix to the list of temperatures to follow, for an autonomous robot - to the coordinates of obstacles in a map, which was rather simple. For the LKAS case study we needed to transform the $TC$ matrix to a set of 2D coordinates, that will produce valid roads after cubic spline approximation. This conversion was more complex and we developed a new technique leveraging affine transformations to vectors.
Next challenge was to create an approximated model. For the autonomous robot we used an implementation provided by the Python Robotics project. For the LKAS, we implemented the model from scratch. The available open source implementations were rather time consuming to execute. We also created the model from the real data for the thermostat case study as we did not find any open source full models.

Finally, it was challenging to find baselines and pipelines to evaluate the produced scenarios. For the autonomous robot, we implemented a simple test evaluation pipeline, based on the Player/Stage simulator. More advanced simulators require the manual creation of the scenarios in the 3D design tools. Fortunately, for the LKAS case study we could use the test evaluation pipeline provided by the SBST2021 competition. 

In conclusion, we advocate for the creation of open source approximated and full models of CPS. Moreover, it is important to establish more test evaluation pipelines and baselines, similar to the LKAS system for other domains of CPS. Finally, we surmise that the CPS simulators should provide a possibility to create environment from configuration files or an API to automate the design of environments.

%Given the same time budget, the single objective configuration of AmbieGen can find the solutions with the higher fitness function value. The multiobjective configuration allows to produce more diverse solutions. To achieve the same fitness as the Ambiegen SO, a bigger number of evaluations is required.

%\textit{Approximated model versus full model for scenario generation.}
%The multiobjective configuration of AmbieGen always produces more diverse solutions.
%In all the considered problems MO produced more diverse test cases with large effect size for thermostat and robot problem and medium for the vehicle problem.

%\textit{Comparing our approach with the available baselines.} 
%Our framework produces the same number of faults as the state-of-the art tool for the LKAS testing. It also outperforms the random research, random generator and the full model based configuration.
%Comparing to the openly available baseline, Frenetic, our approach reveals the similar number of faults given the same time budget. The revealed faults have similar diversity. It outperforms the baseline in the proportion of the valid generated scenarios. Our approach also outperforms the random search in the number of revealed faults and their diversity.
%\section{References
 \section{Threats to validity{\label{sec:threat}}}
%\Foutse{please organize this section following the structure of this paper (Section V): https://web.cs.dal.ca/~masud/papers/masud-ICSME2020-pp.pdf}
We now discuss potential threats to the validity of the results of our study, following existing guidelines \cite{yin2002applications}.

\textbf{Internal validity.} 
To minimize the threats to internal validity, relating to experimental errors and biases, whenever available, we used standardized frameworks for development and evaluation. We implemented all the evolutionary search algorithms (GA and NSGA-II) using a standartd Python Pymoo framework. To evaluate the scenarios for the LKAS case study we used a standardized test pipeline used for SBST2021 workshop tool competition. For\Dima{the}autonomous robot case study we created a customized test evaluation \\pipeline, which is based on the open source Player/Stage robotic simulator. It provides implementations of the widely used robotic models, such as Pioneer 3-AT, and planning algorithms. This simulator was previously used by researchers to conduct similar evaluations, as in \cite{arnold2013testing}.

\textbf{Conclusion validity.}
 Conclusion validity is related to random variations and inappropriate
use of statistics. To mitigate it, we followed the guidelines in \cite{arcuri2014hitchhiker} for search-based algorithm evaluation. We ran each evaluation at least 30 times and ensured the statistical significance of the results by using a two-tailed non-parametric Mann-Whitney U test and Cliff's delta. 
%We presented a detailed description of our
%case studies and the search algorithm. We also provide the code we used \cite{} to facilitate reproducibility.

%We identify two threats to internal validity, that relate to experimental errors and biases. 
%Firstly, the results for performance measurements depend on the fitness function implementation. For simpler problems, random search might outperform the evolutionary search. Secondly, we ran the experiments with a fixed time budget. Bigger time budgets could show difference in the results.
\textbf{Construct validity.} Construct validity is related to the degree to which an evaluation measures what it claims. To compare the test generation algorithms we gave the same time budget to all the algorithms to produce the solutions. For all the algorithms we evaluated the best fitness found, accounting for the scenario fault revealing power. To compare the tools in terms of number of revealed faults we gave each tool the same time budget to produce the scenarios.
To measure the diversity of the test scenarios we used a standard metric such as Jaccard distance, previously used in other studies to compare the difference between the test cases. The exact implementation of this metric is, however, case study specific and thus can introduce some additional bias. Furthermore, the results produced by AmbieGen depend on the implementation of the approximated model. Presumably, higher quality surrogate models can produce more failures of the full model and improve the AmbieGen performance.

\textbf{External validity.} External validity relates to generalizability of our results. We demonstrated how our framework can be applied to generate environments for three different autonomous cyber-physical agents. However, we only considered a limited number of test subjects and limited levels of environment complexity. Therefore more problems should be addressed with different agents and higher environment complexity to make definitive conclusions about the generalizability of AmbieGen.
Nonetheless, our evaluations demonstrated that AmbieGen was effective in revealing unwanted behaviours for all the three considered autonomous CPS agents.
% We only evaluate our approach on three problems, which is not enough to make conclusions about wide application of our technique to different test generation tasks. 
 \section{Conclusions and future work{\label{sec:conc}}}
In this paper we presented the design of AmbieGen, a framework for generating virtual environments for testing autonomous cyber-physical systems. It leverages evolutionary search guided by the approximated model of system. We applied it to generating scenarios for the smart-thermostat, autonomous robotic system, and vehicle lane keeping assist system. Given the same time budget, AmbieGen could generate on average 40 - 50 \% fitter solutions, than random search. Moreover, AmbieGen was effective at detecting faults of the full model. In two hours it could find 9 failures of the Pioneer 3-AT mobile robot in the Player/Stage simulator, comparing to only two failures found by random search. For the full model of the vehicle, equipped with lane keeping assist system, AmbieGen found 14 failures on average, the same as the state of the art baseline - Frenetic. Random search only found 8 failures on average. AmbieGen outperformed Frenetic in the number of valid generated scenarios with a large effect size.

Comparing the two proposed configurations of AmbieGen, the single objective (AmbieGen SO) and multi objective (AmbieGen MO), AmbieGen SO may find fitter solutions than AmbieGen MO given the same time budget. In two hours, for the autonomous robot case study, AmbieGen SO found 10 \% fitter solutions than AmbieGen MO. For the other case studies the difference in the best found solutions fitness was insignificant. Overall, we recommend using the multi objective configuration of AmbieGen, AmbieGen MO, as it always produced a more diverse set of solutions with medium to large effect size and on average could find almost as fit solutions as AmbieGen SO. %converges faster to more fit solutions, than the multi-objective search. AmbieGen MO allows to produce a more diverse set of solutions after running the search. Overall, 

We plan to continue the research in four directions. First is creating more complex environments, taking into account the weather, environmental conditions, and the moving obstacles such as other robots or cars. We also plan to expand the scenario generation to other CPS, such as drone and robot swarms.
Secondly, we will explore the possibility to create more precise simplified models using the system identification techniques, including neural networks and NARIMAX models.
Thirdly, it is important to have  the pipelines for evaluating the generated scenarios. 
We plan to improve our evaluation pipeline for autonomous robots by using more sophisticated simulators such as Argos and Gazebo and more complex models of robots. We will also work on developing evaluation  pipelines dedicated to other types of CPS.  
Finally, we plan to implement AmbieGen as a python framework with an API for generating virtual environments and make it open source.

 \section{Acknowledgements{\label{sec:ack}}}
 This work is partly funded by the Natural Sciences and Engineering Research Council of Canada (NSERC) [Grant No: RGPIN-2019-06956] .
\balance

\bibliographystyle{cas-model2-names}
\bibliography{refs} % Entries are in the
\end{document}